\documentclass[letterpaper, 10 pt, conference]{ieeeconf}  %

\usepackage[utf8]{inputenc}

\IEEEoverridecommandlockouts
\usepackage[american]{babel}
\usepackage{cite}
\usepackage{float}

\usepackage{booktabs}
\usepackage{url}
\usepackage{graphicx}
\usepackage[font=small,labelfont=bf]{caption}
\usepackage{subcaption}
\usepackage{gensymb}
\usepackage{multicol}
\usepackage{array,multirow}
\usepackage{amssymb}
\usepackage{amsmath}
\usepackage{hyperref}
\addto\extrasamerican{%

}
\usepackage{balance}

\usepackage{microtype}

\newcommand*\mean[1]{\overline{#1}}

\title{\LARGE \bf
MICP-L: Mesh-based ICP for Robot Localization using Hardware-Accelerated Ray Casting
}

\author{Alexander Mock$^{1}$, Thomas Wiemann$^{3}$, Sebastian Pütz$^{2}$ and Joachim Hertzberg$^{1, 2}$ %
\thanks{$^{1}$Knowledge-Based Systems group, Institute of Computer Science, Osnabrück University, Hamburger Straße 24, 49084 Osnabrück, Germany {\tt\small amock@uos.de, joachim.hertzberg@uos.de}}%
\thanks{$^{2}$DFKI Robotics Innovation Center, Osnabrück branch, Hamburger Straße 24, 49084 Osnabrück, Germany {\tt\small sebastian.puetz@dfki.de}}%
\thanks{$^{3}$ Robotics in Computer Science, Fulda University of Applied Sciences, Fulda, Germany {\tt\small thomas.wiemann@informatik.hs-fulda.de}}%
\thanks{The DFKI Niedersachsen Lab (DFKI NI) is sponsored by the Ministry of Science and Culture of Lower Saxony and the VolkswagenStiftung}
}

\begin{document}

\setlength{\abovedisplayskip}{3pt}
\setlength{\belowdisplayskip}{3pt}

\maketitle
\thispagestyle{empty}
\pagestyle{empty}

\begin{abstract}
Triangle mesh maps are a versatile 3D environment representation for robots to navigate in challenging indoor and outdoor environments exhibiting tunnels, hills and varying slopes.
To make use of these mesh maps, methods are needed to accurately localize robots in such maps to perform essential tasks like path planning and navigation. 
We present \textit{Mesh ICP Localization} (MICP-L), a novel and computationally efficient method for registering one or more range sensors to a triangle mesh map to continuously localize a robot in 6D, even in GPS-denied environments.
We accelerate the computation of \textit{ray casting correspondences} (RCC) between range sensors and mesh maps by supporting different parallel computing devices like multicore CPUs, GPUs and the latest NVIDIA RTX hardware.
By additionally transforming the covariance computation into a reduction operation, we can optimize the initial guessed poses in parallel on CPUs or GPUs, making our implementation applicable in real-time on many architectures.
We demonstrate the robustness of our localization approach with datasets from agricultural, aerial, and automotive domains.
\end{abstract}

\section{Introduction}

Localization is the task of estimating the state of a mobile robot in a reference coordinate system. 
It is the foundation for any mobile robot to operate autonomously in a given application environment.
In many outdoor scenarios, localization is done via GPS or triangulation with predefined landmarks.
However, in many cases, GPS is unreliable or not available at all, therefore other localization strategies have to be considered.
Algorithms such as LiDAR-based SLAM have been developed to address this problem.
In recent years, such SLAM algorithms have been improved in a way that even robots with low computational resources can estimate their current state in 6DoF while simultaneously mapping their environment in 3D.
An increasingly important map representation in the context of SLAM is a Truncated Signed Distance Field (TSDF)~\cite{meisoldt21ecmr}.
TSDFs can be transferred easily into 3D triangle meshes to compress the information encoded in the original map.
Recently, algorithms have been developed to plan the motion of a robot over the mesh's surface to a given goal~\cite{puetz2021cvp}.
To execute such plans, reliable, frequent, and accurate localization in mesh maps is required. 

\begin{figure}[t]
\centering
\vspace{0.3cm}
\includegraphics[trim=0 60 0 10, clip, width=0.98\linewidth]{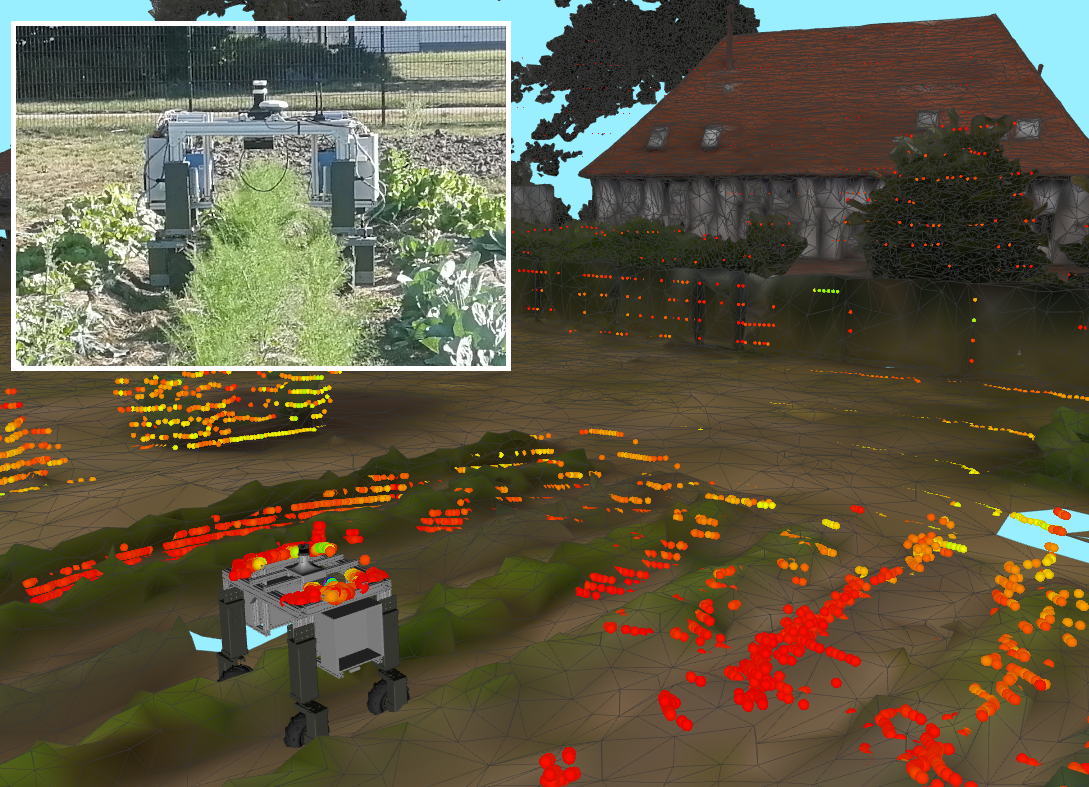}
\caption{The Lero agricultural monitoring robot uses MICP-L for localization for mesh-based navigation between beds in a market garden micro-farming environment.}
\label{fig:teaser}
\vspace{-0.5cm}
\end{figure}

Our paper presents a novel approach called MICP-L\footnote{Source code is available here~\url{https://github.com/uos/rmcl}}, which enables robots equipped with arbitrary range sensors to be localized directly in triangle mesh maps. 
It is designed to be applicable to robots with varying computational capabilities. 
For that, MICP-L builds upon the open-source library Rmagine, presented in previous work~\cite{mock2023rmagine}.
We contribute
\begin{itemize}
\item Robust range-sensor-to-mesh registration via hardware-accelerated ray casting correspondences (RCC),
\item Accurate and reliable localization, tested in various real-world domains as shown in~\autoref{fig:teaser},
\item Flexible workload distribution over CPU or GPU, enabling application on various hardware platforms,
\item Support of arbitrary range sensor combinations.
\end{itemize}

\section{Related Work}

Localization is a key capability required in many robotics applications and in the last decade researchers generated many approaches using various map representations, ranging from point clouds~\cite{shan2018lego, liosam2020shan, fastlio2021, fasterlio2022, vizzo2023} to occupancy grids, signed distance fields (SDF)~\cite{oleynikova2017voxblox, vizzo2022sensors, meisoldt21ecmr}, meshes~\cite{vizzo2021puma, dai2017bundlefusion, akai2020, dreher2021global, chen2021range, ruan2023slamesh}, and more, e.g., neural representations~\cite{zhong2023, wiesmann2023}.

Generating 3D triangle meshes online~\cite{ruan2023slamesh}, on system-on-chip (SoC) devices~\cite{meisoldt21ecmr} and for large scales environments~\cite{wiemann2018irc, dai2017bundlefusion, vizzo2021puma, lin2023immesh} have become available.
Meshes feature a closed surface and can be used for planning and executing paths of autonomously driving robots in unstructured environments~\cite{puetz2021cvp}. 
The success of mesh navigation software heavily depends on the robot's ability to frequently and reliably localize itself with respect to the mesh map.
In principle, global localization in triangle meshes can be done using particle filters~\cite{dreher2021global}.
Chen et al. introduced a Monte Carlo Localization (MCL) method for meshes, which utilizes LiDARs and OpenGL-simulated range images for importance sampling~\cite{chen2021range}.
However, particle filter-based approaches produce high computational loads to track the robot's pose. 
To overcome this issue, relative localization methods such as iterative closest point (ICP) are often used to track the robot's pose~\cite{rusinkiewicz2001efficient, koide2021vgicp, shan2018lego, liosam2020shan, segal2009gicp}.
All of them register a measured point cloud to a reference point cloud.
Several publications, spanning from computer graphics to robotics research, have presented ICP-like methods that can register point clouds to reference meshes instead~\cite{holz2014registration, li2015modified, avetisyan2019scan2cad, mejia2019pcl2mesh, bourquat2022hierarchical, vizzo2021puma}.

In 2019, Mejia-Parra et al. solved point cloud to mesh registration with a novel approach of finding the closest surface point for each point cloud point using hash grids~\cite{mejia2019pcl2mesh}.
They evaluated their approach by registering point clouds to single meshes of closed objects.
However, dealing with meshes of large environments, as in~\cite{chen2021range}, introduces challenges beyond the scope of their methodology.
For example, triangle mesh maps of interiors usually consist of walls with two parallel surfaces. 
If the initial pose estimate deviates just a little from the optimum, this would lead to matches between wrong wall surfaces.
In addition, large meshes would drastically increase the size of the hash table, making it impossible to use on mobile robots.
In 2021, Vizzo et al. presented an approach for mesh-based SLAM named PUMA~\cite{vizzo2021puma}.
They generated a mesh from a scan and registered the following scan to it by using ray casting correspondences (RCC).
Their correspondences seem to be suitable to solve the mentioned issues of~\cite{mejia2019pcl2mesh}.
However, real-time capability was not even achieved for fractions of the entire mesh map, making it unsuitable for online operation on autonomous vehicles.

With MICP-L, we address the problems of~\cite{vizzo2021puma} and~\cite{mejia2019pcl2mesh} by introducing a light-weight method to find RC correspondences and optimize the pose estimation by utilizing NVIDIA's RTX hardware for acceleration.
We reach more than real-time localization even in large-scale mesh maps and overcome convergence issues of classical point cloud-based ICP~\cite{zhang2022icp}.

\section{MICP-L}

MICP-L continuously registers range data acquired from one or multiple range sensors to a triangle mesh map, starting at an initial pose estimate.
The general workflow can be summarized as follows:

\begin{enumerate}
\item Find ray casting (RC) or closest point (CP) correspondences,
\item Estimate the optimal transformation parameters $\Delta T$ through covariance reduction and SVD using either P2P or P2L as error measure, 
\item Apply $\Delta T$ to the pose guess and repeat step 1).
\end{enumerate}

\subsection{Closest Point Correspondences}

For point-cloud maps, locating the nearest model point to a dataset point is typically sped up by pre-building kd-trees or hash grids over the model.
In mesh maps, the task differs slightly as finding the closest point on a mesh surface may not necessarily involve vertices but requires searching the nearest point on each face.
We use the BVH implementation of Embree~\cite{embree2014} to accelerate this operation.

\subsection{Ray Casting Correspondences}

\begin{figure}
\centering
\includegraphics[trim=0 170 0 130,clip,width=0.99\linewidth]{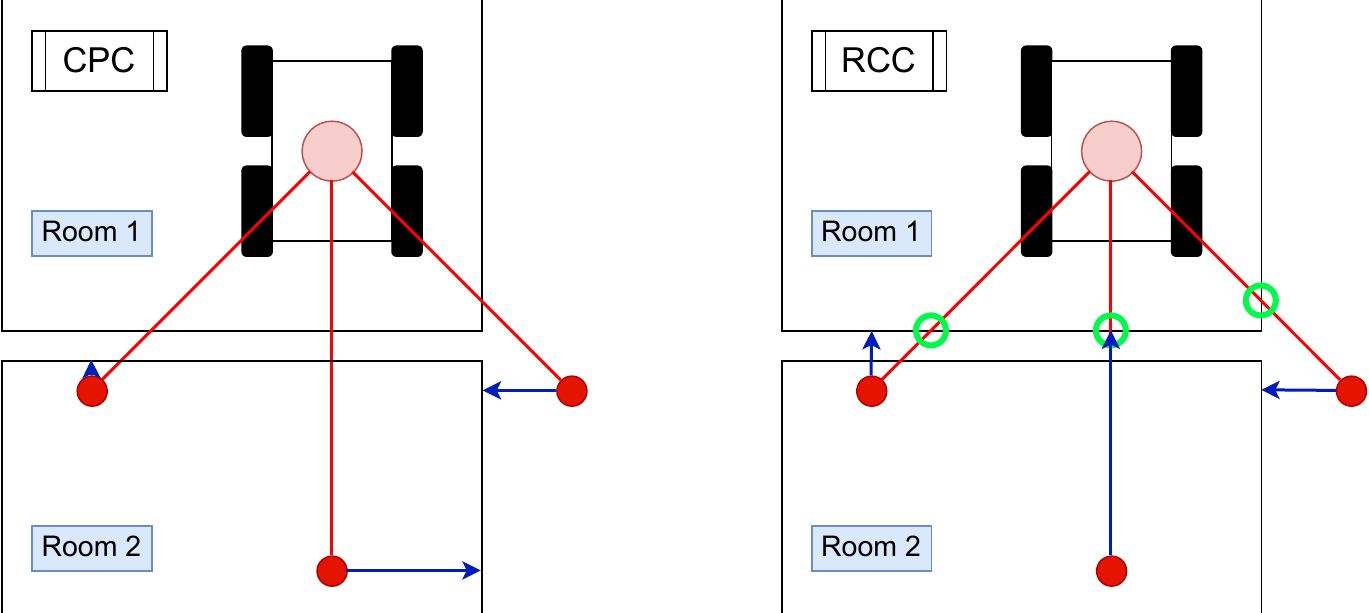}
\caption{A 2D map consists of two rooms.
Left: Conventional closest point correspondences fail since they match on the wall's surface of the wrong room.
Right: RCCs overcome these problems, by first ray casting along the real measurement's direction (red line) to find the next intersection with the map (green circle).
For point-to-plane (P2L) optimization, the actual correspondence is the projection onto the plane (dark blue arrow).}
\label{fig:corr}
\vspace{-0.5cm}
\end{figure}

Dealing with meshes allows us to use specialized RTX chips on the NVidia GPUs to efficiently compute ray-to-mesh intersections.
We use this technique to accelerate finding so-called ray casting correspondences~(RCC), introduced with PUMA~\cite{vizzo2021puma}.
For computing RCCs, we first convert each measurement $i$ of a range sensor into a ray representation, consisting of origin, direction, and range.
Next, we use the given pose estimate to trace virtual rays along the paths that the real sensor would scan.
The intersection with the map represents the actual point that would be measured if the sensor were at the estimated pose.
We then project the Cartesian point of the real scan measurement onto the plane of intersection to determine the map correspondence.
These RCCs allow matching scan points only to the surface closest to the range sensor, as shown in~\autoref{fig:corr}.
This reduces the likelihood of drifting into other rooms during registration as shown in the experiments, subsequently.

We use Rmagine, a library to flexibly build ray tracing-based sensor simulations~\cite{mock2023rmagine}, for all ray casts.
It decides at run time whether to compute the RCC on CPU~\cite{embree2014} or on a NVIDIA GPU and, if available, utilize the RTX units for hardware-acceleration~\cite{optix2010}.

\subsection{Correction}

Finding RC or CP correspondences returns a list of scan points $D$ and map points $M$, where the $i$-th scan point $d_i$ corresponds to the $i$-th map point $m_i$.
All points are located in the same coordinate system, e.g., the robots base frame.

Next, we search for the transformation parameters $\Delta T$, consisting of a rotation $\Delta R$ and a translation $\Delta t$, that best fit the scan points (dataset) $D$ to the map points (model) $M$.
This can be done by minimizing the following equation:
\begin{equation}
    E(\Delta R, \Delta t) = \frac{1}{n} \sum_{i=1}^{n} \| m_i - (\Delta R d_i + \Delta t) \|^2_2
\end{equation}
Our solver is based on Umeyama's method~\cite{umeyama1991} and enhances the computation of the covariance matrix for both CPU and GPU by transforming it into a reduction operation, which is a standard pattern for parallel computing.

\begin{figure}[t]
\centering
\vspace{0.15cm}
\includegraphics[width=0.92\linewidth]{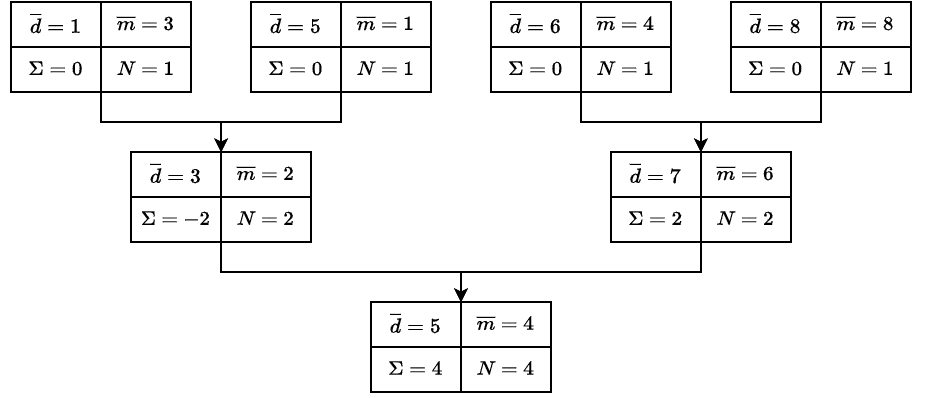}
\caption{1D example of covariance reduction given 4 correspondences utilizing Eq.~(\ref{eq:cov:1}-\ref{eq:cov:3}). }
\label{fig:covred}
\vspace{-0.5cm}
\end{figure}

Given two arbitrary partitions of correspondences A and B that are used to perform the registration operations $D_A \rightarrow M_A$ and $D_B \rightarrow M_B$.
Then $A$ has the partition parameters $\left(\mean{m}_A, \mean{d}_A, \Sigma_A, N_A\right)$ and $B$ has the partition parameters $\left(\mean{m}_B, \mean{d}_B, \Sigma_B, N_B \right)$.
Given those parameters, we search for the parameters $\left(\mean{m},\mean{d},\Sigma,N\right)$ of the concatenated correspondences by only using the given partition's parameters.
We begin with updating the total number of measurements $n$, dataset and model means $\mean{m}$, $\mean{d}$ by 
\begin{align}\label{eq:cov:1}
    N = N_A + N_B && c_A = N_A/N && c_B = N_B / N
\end{align}
\vspace{-0.5cm}
\begin{align}\label{eq:cov:2}
    \mean{d} = c_A \cdot \mean{d}_A + c_B \cdot \mean{d}_B && \mean{m} = c_A \cdot \mean{m}_A + c_B \cdot \mean{m}_B\text{,}
\end{align}%
and then merging the covariances to $\Sigma$ by

\noindent\begin{equation}\label{eq:cov:3}
\begin{aligned}
\Sigma &= c_A \cdot \left(\Sigma_A + \left(\mean{m}_A - \mean{m}\right)\left(\mean{d}_A - \mean{d}\right)^T \right) \\
& + c_B \cdot \left( \Sigma_B + \left(\mean{m}_B - \mean{m}\right)\left(\mean{d}_B - \mean{d}\right)^T \right)\text{.}
\end{aligned}
\end{equation}%
From these more general equations, the conventional Umeyama equations can be reconstructed by setting the partition $A$ as the current collection of all correspondences including their mean and covariances beginning with $\left(\mean{d}_A, \mean{m}_A, \Sigma_A, N_A\right) = \left(0,0,0,0\right)$.
For every new correspondence $\left(d_i, m_i\right)$ we build a new partition $B$ with the parameters $\left(\mean{d}_B, \mean{m}_B, \Sigma_B, N_B\right) = \left(d_i, m_i, 0, 1\right)$.
After applying Eq.~(\ref{eq:cov:1}-\ref{eq:cov:3}), we override the partition parameters of $A$ with the results and repeat the whole procedure for each new correspondence pair.
This gives an exact iterative version analog to the regular two-pass covariance computation method used in~\cite{umeyama1991}.
Furthermore, Eq.~(\ref{eq:cov:1}-\ref{eq:cov:3}) enable us to create well-balanced CUDA kernels using reductions, as demonstrated in~\autoref{fig:covred}.
By applying this to our problem, we can reduce the RCC to two means, $\mean{d}$ and $\mean{m}$, and one covariance matrix $\Sigma$ per sensor.
In our implementation, this is done either on the CPU with OpenMP or on the GPU using CUDA kernels.

\begin{figure}[t]
    \centering
    \includegraphics[width=0.98\linewidth]{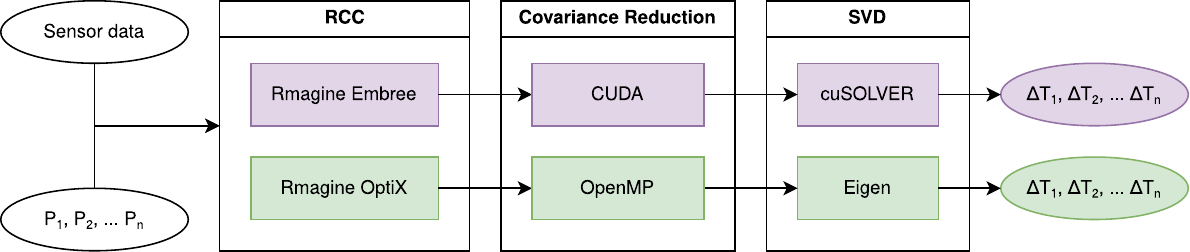}
    \caption{Workflow of MICP-L Correctors. CPU and GPU computations are colored in green and purple respectively. Both the GPU and the CPU implementation receive a list of pose estimates and sensor data. The outputs are delta transforms for each input pose.}
    \label{fig:workflow}
    \vspace{-0.5cm}
\end{figure}

If the robot has only one range sensor, we continue with computing the correction parameters $\Delta R$ and $\Delta t$ through singular value decomposition (SVD) either using Eigen or CUDA's cuSOLVER according to the original procedure of Umeyama.
Depending on whether the calculation is done on the CPU or the GPU, the inputs and outputs are located in CUDA buffers or in the RAM of the CPU, respectively.
This allows the development of pre- and post-processing steps, where data is transferred between devices only on demand to ensure maximum performance.
After correcting the pose guess, we repeat the steps of RCC and correction until convergence.
The entire workflow of both the CPU and the GPU correction is shown in~\autoref{fig:workflow}.

\subsection{Combined Correction}

Using Umeyama's approach and our modifications, it is feasible to combine multiple range sensors and compute a unified correction.
For each sensor $j$, we compute the means $\mean{d}_j$, $\mean{m}_j$, and covariances $\Sigma_j$.
Since all correspondences are transformed into a shared coordinate system, we can merge the means and covariances of all sensors with Eq.~(\ref{eq:cov:1}-\ref{eq:cov:3}).
After merging the covariances of individual sensors, we can determine the transformation components to obtain a total correction $\Delta T$ that contains the influence of every sensor.

Furthermore, we offer the option to replace the automatic calculation of $c_j$ with user-defined weights to model the impact of certain sensors during correction.
If every sensor measurement is equally reliable, we propose selecting the weights based on the number of correspondences, as described in Eq.~(\ref{eq:cov:1}).
A sensor with many measurements will automatically have a higher weight assigned to it than a sensor with fewer measurements.

\begin{figure}[b]
    \vspace{-0.2cm}
    \centering
    \begin{subfigure}[t]{0.49\linewidth}
        \includegraphics[trim=0 70 0 150,clip,width=\linewidth]{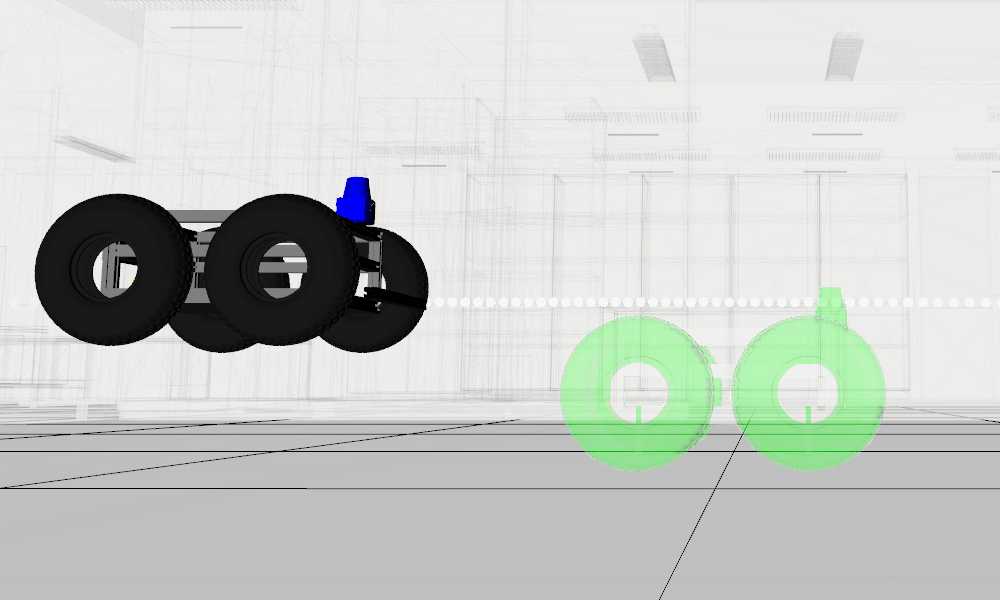}
        \caption{Pose estimated $(-0.5, 0, 0.2)$ off the real pose. }
        \label{fig:ex2d:a}
    \end{subfigure}
    \begin{subfigure}[t]{0.49\linewidth}
        \includegraphics[trim=0 70 0 150,clip,width=\linewidth]{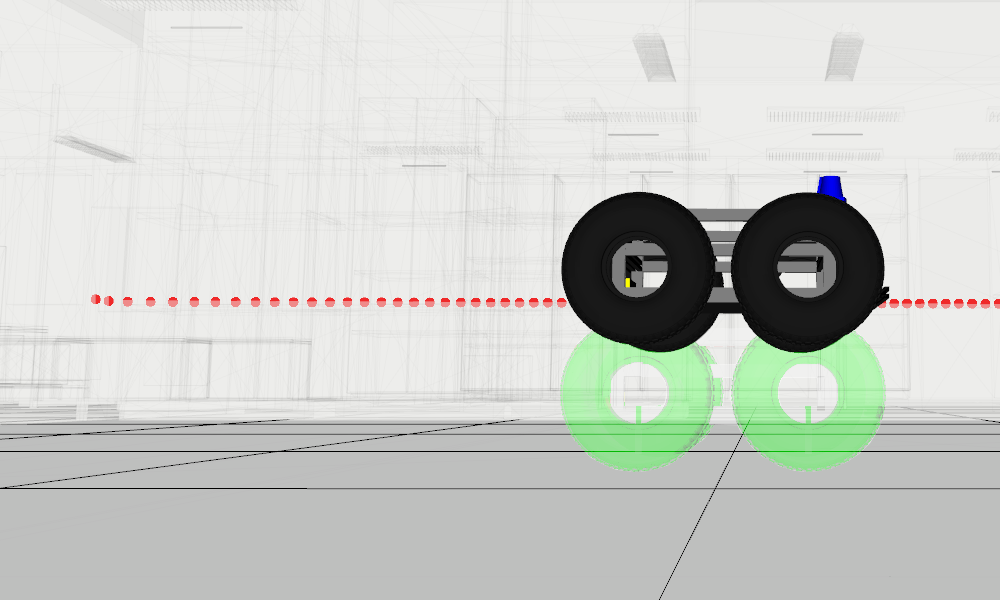}
        \caption{Correction with 2D LiDAR (red) only.}
        \label{fig:ex2d:b}
    \end{subfigure} \\
    \begin{subfigure}[t]{0.49\linewidth}
        \includegraphics[trim=0 70 0 150,clip,width=\linewidth]{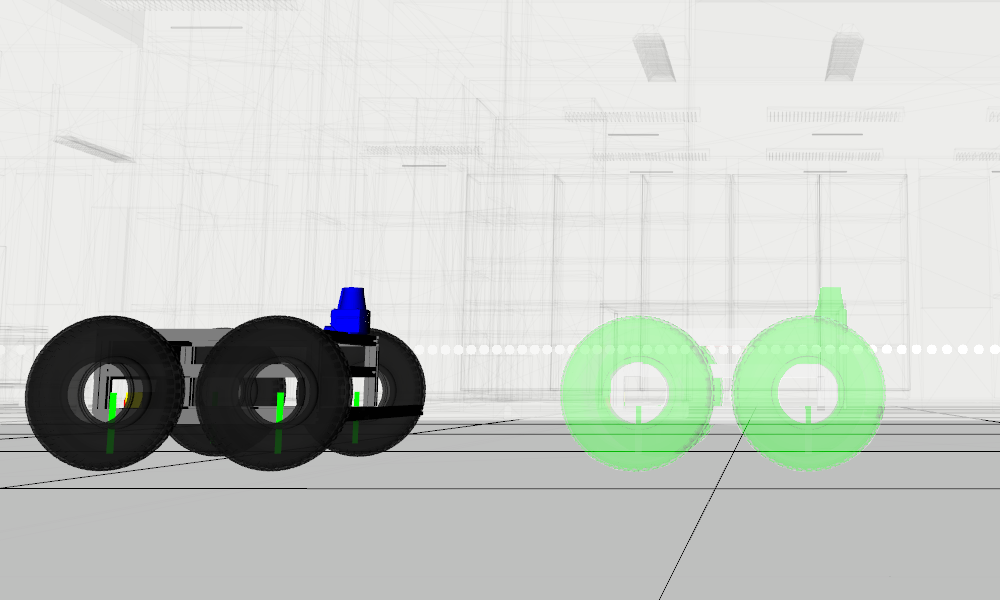} 
        \caption{Correction with virtual wheel sensor (green) only.}
        \label{fig:ex2d:c}
    \end{subfigure}
    \begin{subfigure}[t]{0.49\linewidth}
         \includegraphics[trim=0 70 0 150,clip,width=\linewidth]{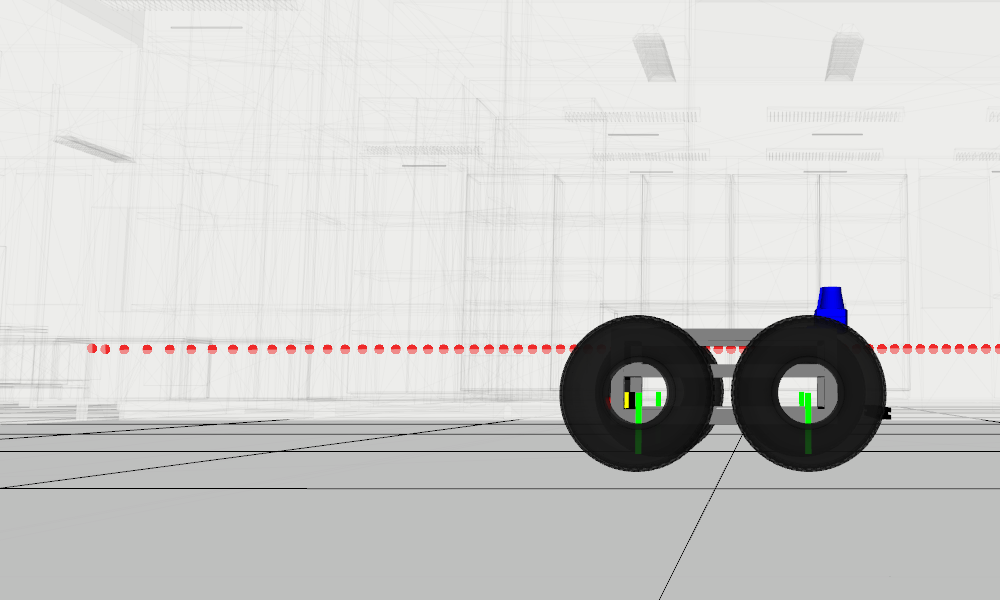}
         \caption{Combined correction of virtual wheels and 2D LiDAR.}
         \label{fig:ex2d:d}
    \end{subfigure}
    \caption{Visualization of a combined correction until convergence. 
    a) shows the pose guess from which the correction starts. 
    It is approximately 0.2\,m over the ground (z) and -0.5\,m in the x direction of the ground truth pose (green robot). 
    b) shows the correction results using only a 2D LiDAR. Here, the robot cannot correct the z error. 
    d) shows the combined correction results considering the single corrections of b) and c) by combining two covariances.}
    \label{fig:ex2d}
\end{figure}

Thus, we enable the correction of multiple sensor sources by merging the intermediate covariances of each sensor.
In our implementation, we provide functions to generate a list of partition parameters (one partition per sensor).
Additionally, we provide functions to fuse the partitions into one, either uniformly weighted considering the number of measurements or completely user-defined by freely choosing the weights for each sensor.
To better understand how they can be combined for correction, consider the following example:
We want to localize a robot in a 3D map. 
This robot is only equipped with a 2D LiDAR, making it difficult to estimate its exact 6DoF pose in the map.
However, we know that our robot is on the ground most of the time.
We can use this information to attach an additional virtual scanner as Rmagine's OnDn model to the robot that permanently scans from the wheel centers to the ground.
This virtual scanner generates fixed distances equal to the wheel radius.
Using this virtual scanner and the 2D LiDAR measurements, we can build two covariance matrices for a given pose estimate.
The combination of those two is visualized in~\autoref{fig:ex2d}.
Thus, MICP-L can localize a robot equipped with various range sensors.
A list of supported configurations is listed and discussed in~\cite{mock2023rmagine}.

\section{Experiments}

\begin{figure}[t]
    \centering
    \includegraphics[trim=0 100 0 100,clip,width=0.98\linewidth]{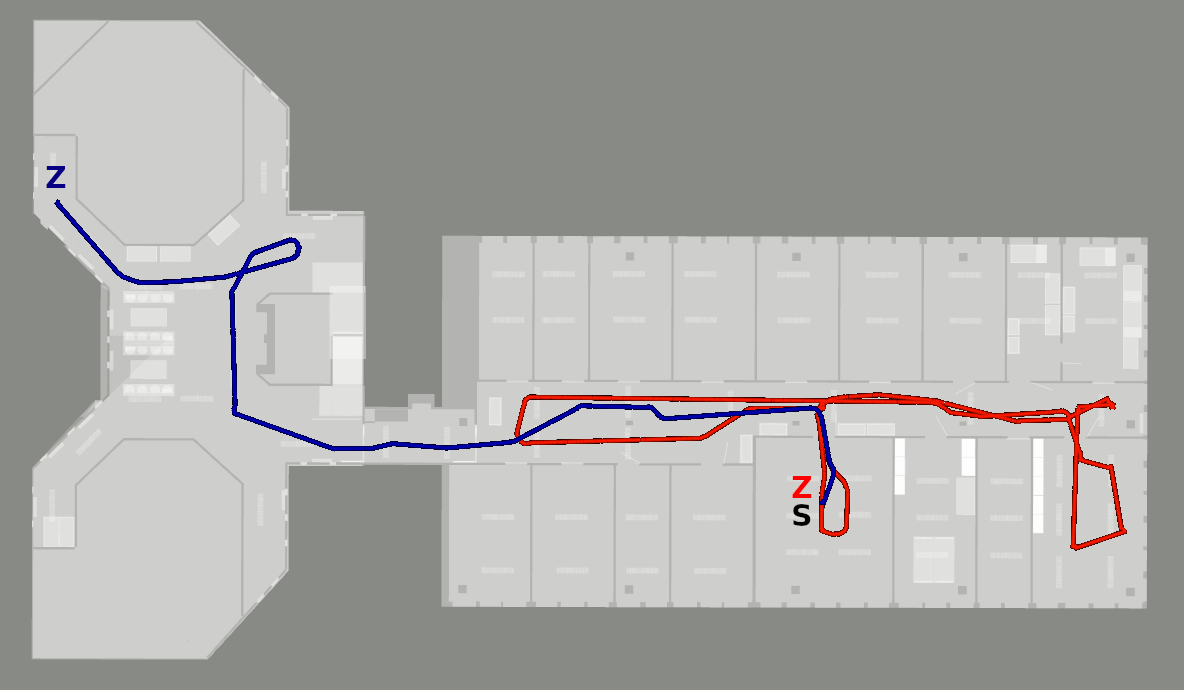}
    \caption{Two test drives in AVZ Gazebo world. The ground truth trajectories are colored in red and blue.}
    \label{fig:gazebo}
    \vspace{-0.5cm}
\end{figure}

Our experiments aim to analyze our software according to our predefined requirements, as well as to provide a comprehensive overview of its capabilities and limitations.
First, we evaluate the localization performance on three challenging real-world domains: agriculture, automotive, and drones.
Second, we compare our hardware-accelerated RCC with classic point-to-plane correspondences.
Third, we give a detailed overview of MICP-L's computational demands.

\subsection{Localization Performance}

Quantifying localization performance necessitates superior localization systems capable of producing precise ground truth data.
To mitigate potential errors associated with ground truth estimation, we initially present results obtained from a simulation environment where perfect localization is inherently guaranteed.
However, simulations have inherent limitations in terms of realism.
Therefore, our second series of experiments focuses on evaluating localization accuracy in real-world datasets.

\subsubsection{Simulation}

For our experiments, we used a Gazebo world called AVZ, available in the ROS package~\texttt{uos\_tools}.
In this world, we placed a robot equipped with a Velodyne VLP-16 and applied Gaussian noise to the simulated LiDAR data with a standard deviation of 0.8\,cm.
First, we run MICP-L giving a stationary robot a pose guess near its real position.
It converged to a translation error of 0.2\,mm.
The fact that the error is smaller than the sensor inaccuracy can be justified by the fact that the individual errors average out on the surfaces during registration.
In the next experiment, the robot moves through the simulated AVZ environment with a simple, but inaccurate pose estimation based on odometry.
MICP-L continuously corrects this estimate to the mesh, and we compared the resulting corrected trajectory to the perfect ground truth provided by Gazebo.
For two test drives, the ground truth trajectories, red and blue, are shown in~\autoref{fig:gazebo}.
As a localization metric, we tracked the mean absolute pose error (APE) to the ground truth trajectory.
The odometry resulted in a mean APE of 8.228\,m and 4.5485\,m for the red and blue trajectory, respectively.
By continuously correcting the odometry using MICP-L, we improved it to 0.98\,cm for the red and 0.86\,cm for the blue trajectory.
This shows that MICP-L is able to correct the inaccurate odometry estimate, achieving an average accuracy of below 1\,cm.

\subsubsection{Real-world}

To demonstrate the performance of MICP-L in real-world scenarios, we tested our software across various domains, each one providing unique challenges.
In agriculture, the environment is particularly unstructured, e.g., parts of the map are constantly changing, like trees moving in the wind or plants growing.
The automotive domain includes driving cars with fast translational changes in large environments.
The drone datasets contain full 6DoF movements, in contrast to the previous domains with fewer degrees of freedom.
Additionally, we selected the drone dataset to specifically assess MICP-L's performance in GPS-denied environments.

\begin{figure}[t]
\centering
\begin{subfigure}[t]{0.99\linewidth}
    \begin{minipage}{0.32\linewidth}
        \centering
        EKF
    \end{minipage}
    \begin{minipage}{0.32\linewidth}
        \centering
        MICP-L
    \end{minipage}
    \begin{minipage}{0.32\linewidth}
        \centering
        RTK-GPS
    \end{minipage}
\end{subfigure}\\
\vspace{0.1cm}
\begin{subfigure}[t]{0.99\linewidth}
    \includegraphics[width=0.32\linewidth]{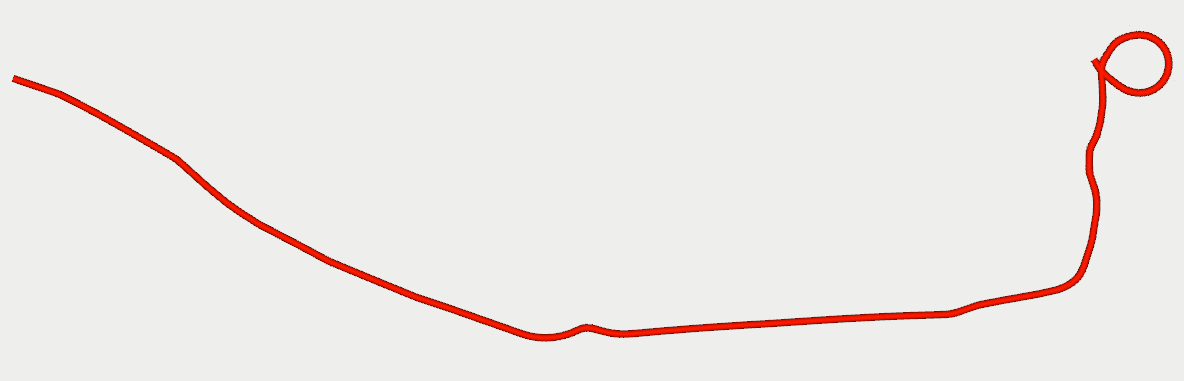}
    \includegraphics[width=0.32\linewidth]{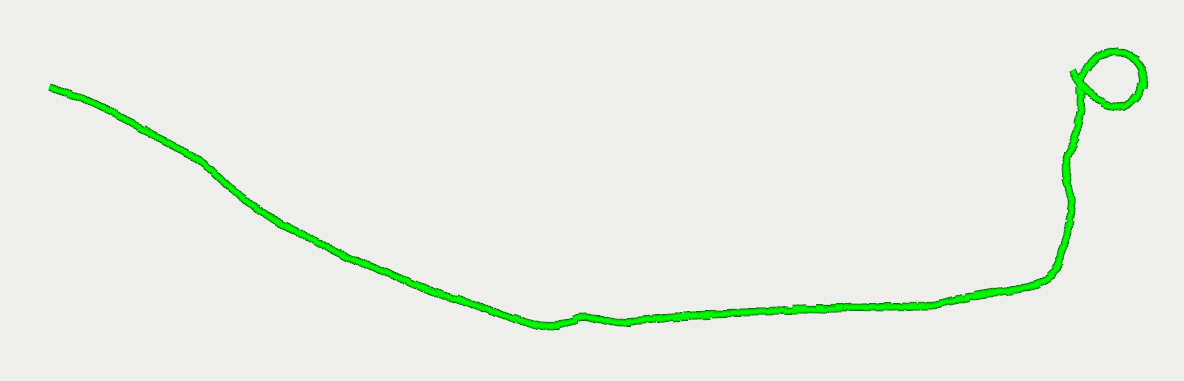}
    \includegraphics[width=0.32\linewidth]{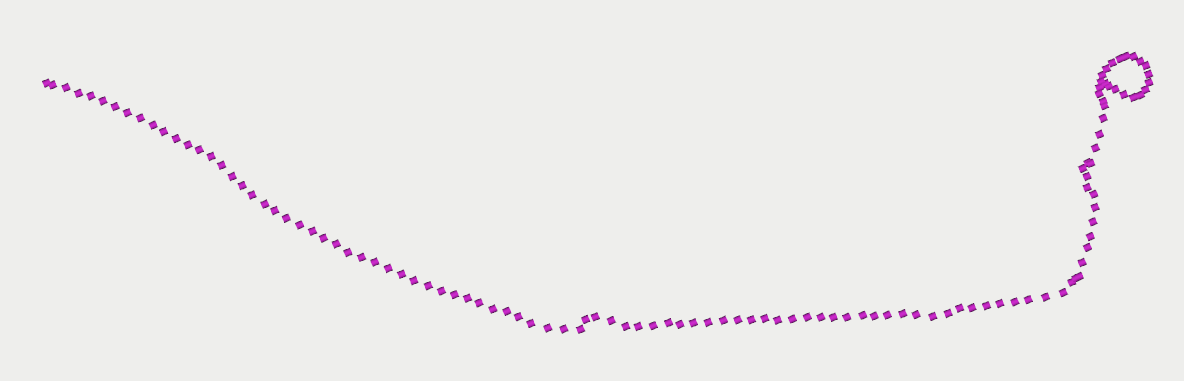}
    \caption{Drive 1: To the field (64 m, 0.32 m/s)}
    \label{fig:lerobags:a}
\end{subfigure}\\
\begin{subfigure}[t]{0.99\linewidth}
    \includegraphics[width=0.32\linewidth]{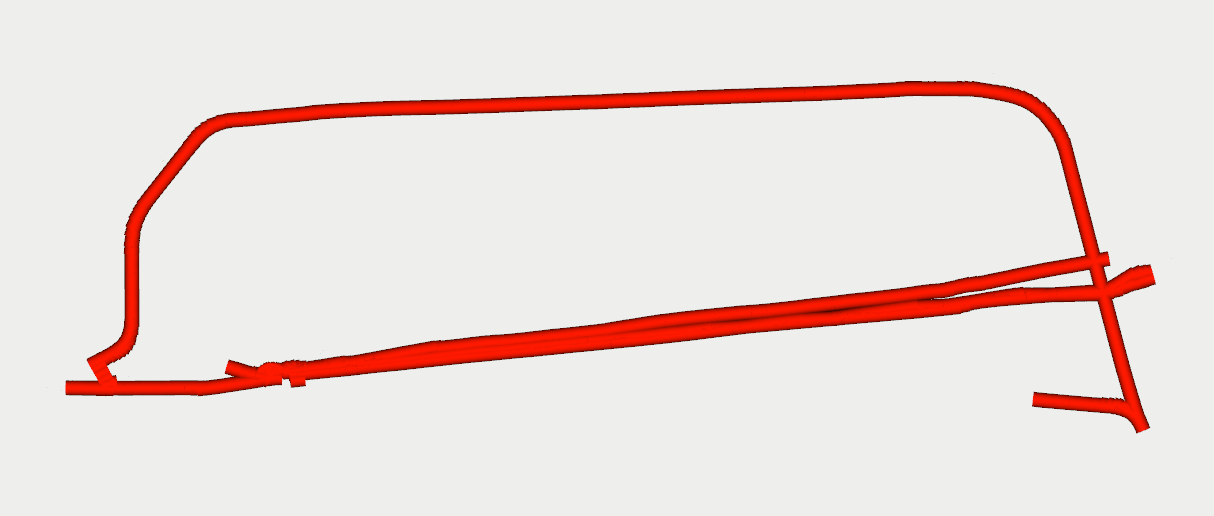}
    \includegraphics[width=0.32\linewidth]{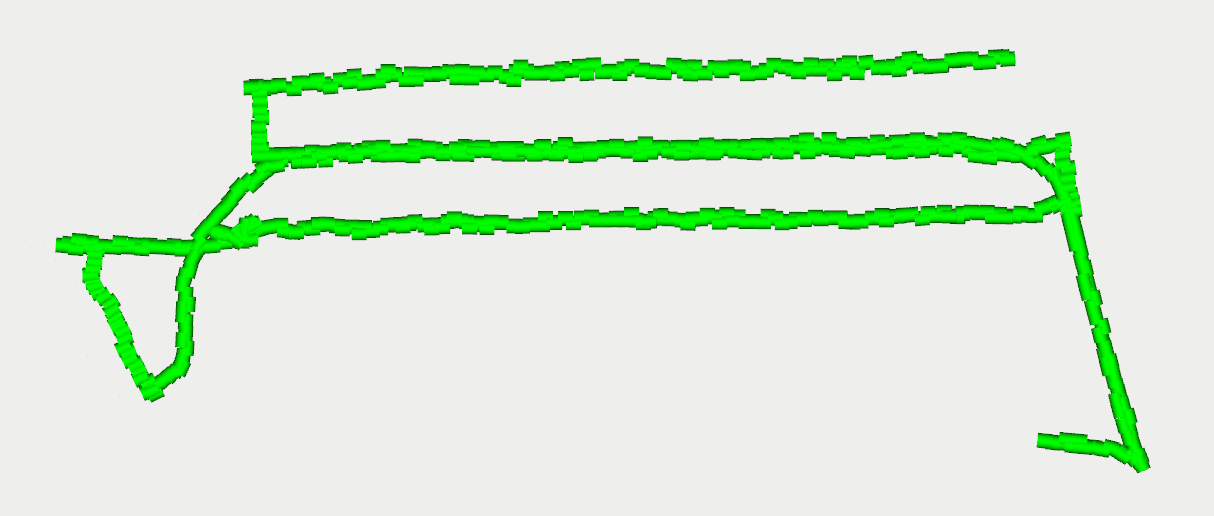}
    \includegraphics[width=0.32\linewidth]{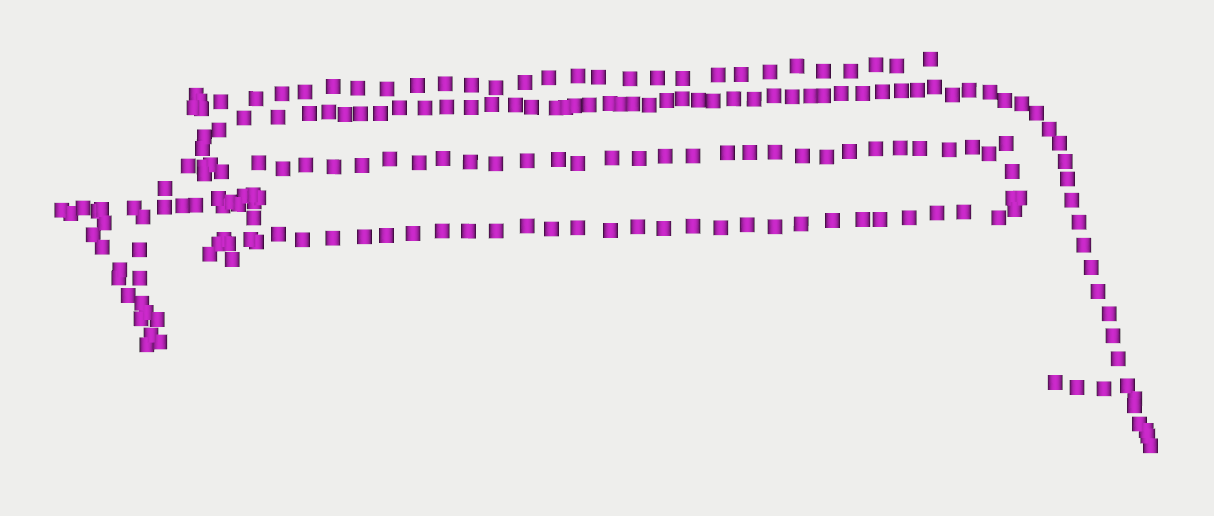}
    \caption{Drive 2: Infield navigation (121 m, 0.48 m/s)}
    \label{fig:lerobags:b}
\end{subfigure}\\
\begin{subfigure}[t]{0.99\linewidth}
    \includegraphics[width=0.32\linewidth]{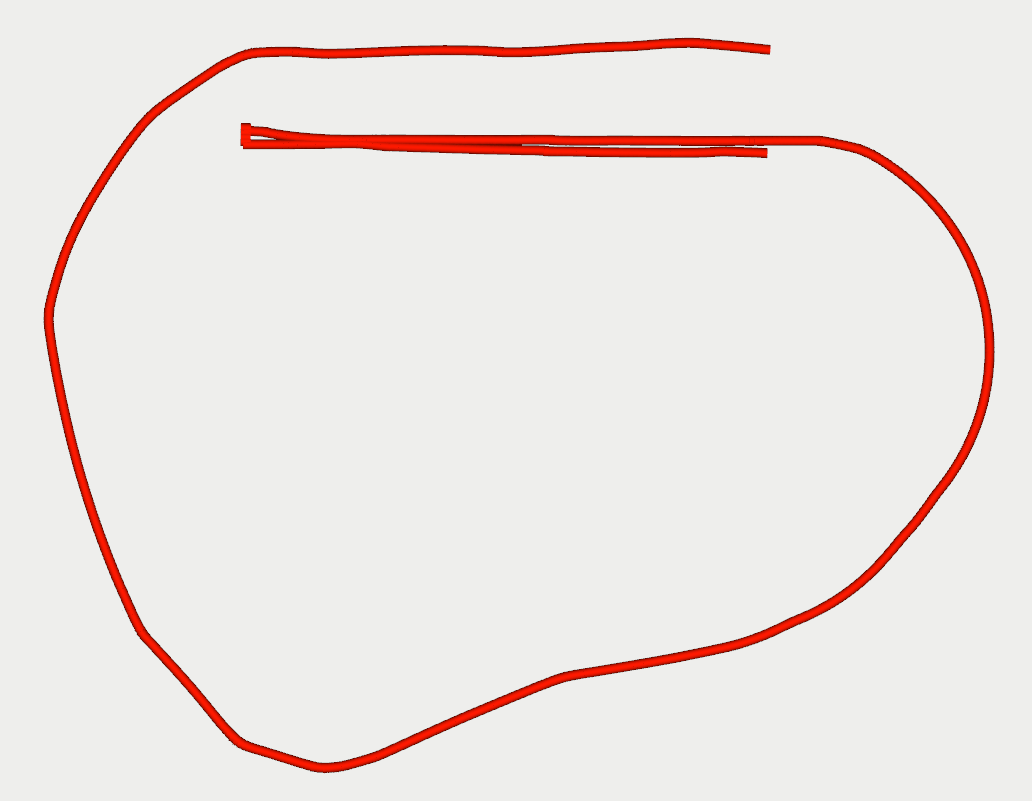}
    \includegraphics[width=0.32\linewidth]{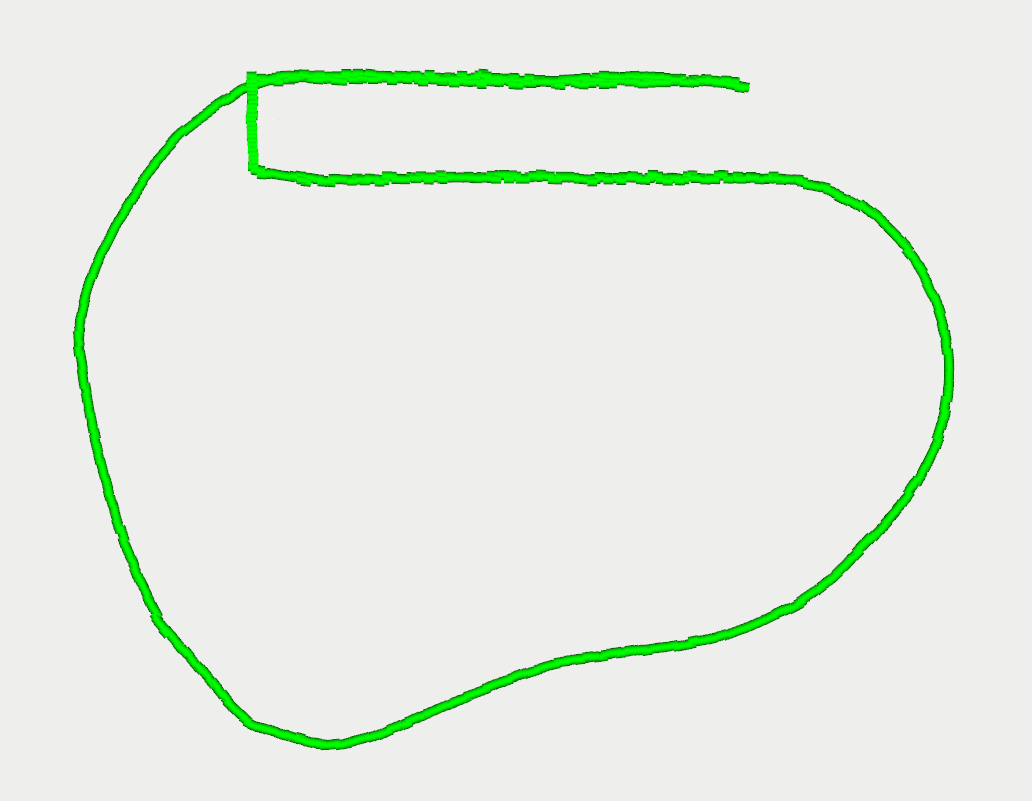}
    \includegraphics[width=0.32\linewidth]{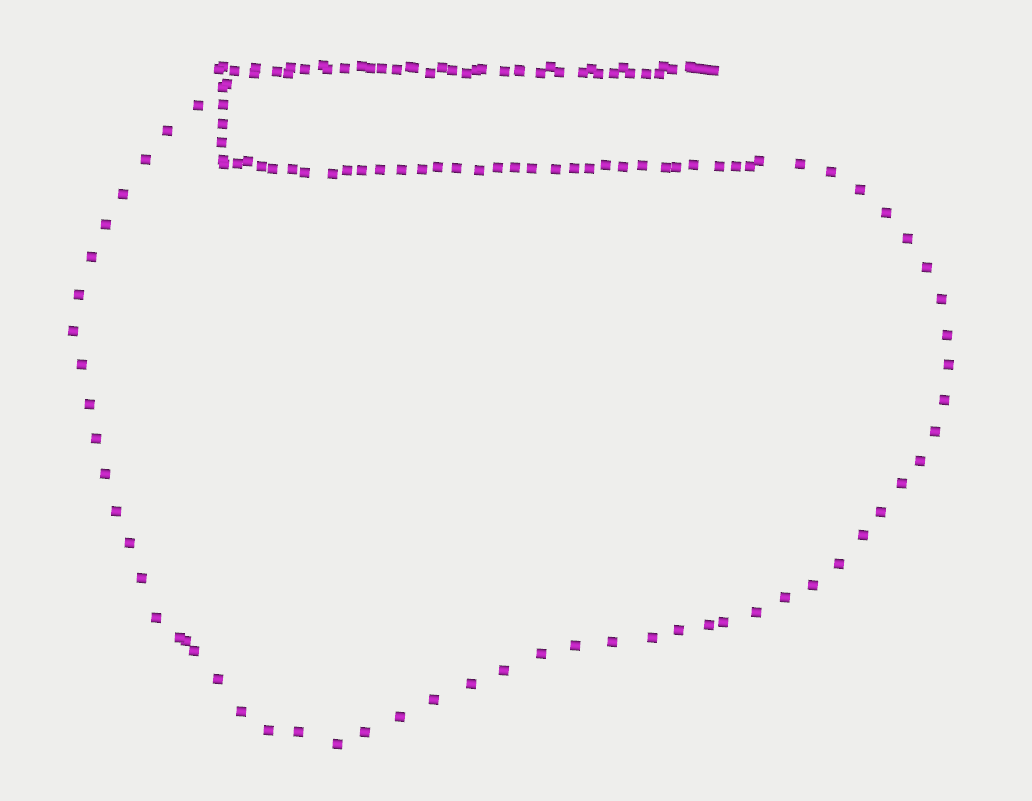}
    \caption{Drive 3: Infield and outfield navigation (107\,m, 0.62\,m/s)}
    \label{fig:lerobags:c}
\end{subfigure}
\caption{Autonomously navigating the agricultural monitoring robot Lero on a field. For each drive a), b), and c) we recorded the trajectory of the EKF~(red), MICP-L~(green), and an RTK-GPS~(purple).}
\label{fig:lerobags}
\vspace{-0.5cm}
\end{figure}

\textbf{LERO - Agricultural:}
The following experiments investigate MICP-L's performance in agricultural environments and the integration in the mesh navigation framework~\cite{puetz2021cvp}. %
For that, we used the agricultural monitoring robot Lero, which is equipped with an Ouster OS1-64 LiDAR.
Lero provides two Intel NUCs, both capable of hardware accelerating MICP-L with RTX GPUs as specified in~\autoref{tab:devices}.
To build a reference map, we recorded a point cloud of the environment using a Z+F IMAGER 5016 high-resolution terrestrial laser scanner and reconstructed a triangle mesh with LVR2~\cite{wiemann2018irc}.
The resulting mesh and the robot are shown in~\autoref{fig:teaser}.
The map consists of around 2 million triangles and covers an area of 154\,m $\times$ 163\,m.
It exhibits many beds with a width of about 75\,cm, usually 10\,m long, and a track width of one meter next to each other.
For autonomous robot navigation, we used the Continuous Vector Field Planner~(CVP)~\cite{puetz2021cvp} and the high-level control framework Move Base Flex~\cite{puetz2018mbf} and recorded three test drives.
We observed that in every test drive, Lero was able to robustly navigate across the field and arrive at the given goals precisely, showing that MICP-L synergizes with the CVP.
Lastly, we extracted the trajectories estimated by MICP-L, the EKF, and an RTK-GPS module, shown in~\autoref{fig:lerobags}.

\begin{figure*}[t] %
    \centering
    \begin{subfigure}{1.0\linewidth}
        \centering
        \includegraphics[trim={0 0 0 0},clip,width=0.325\linewidth]{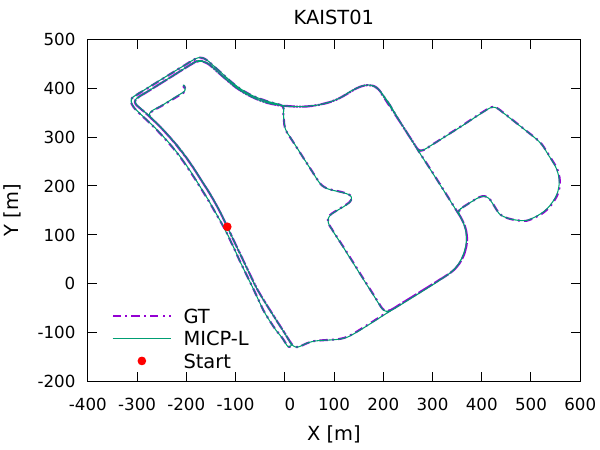}
        \includegraphics[trim={0 0 0 0},clip,width=0.325\linewidth]{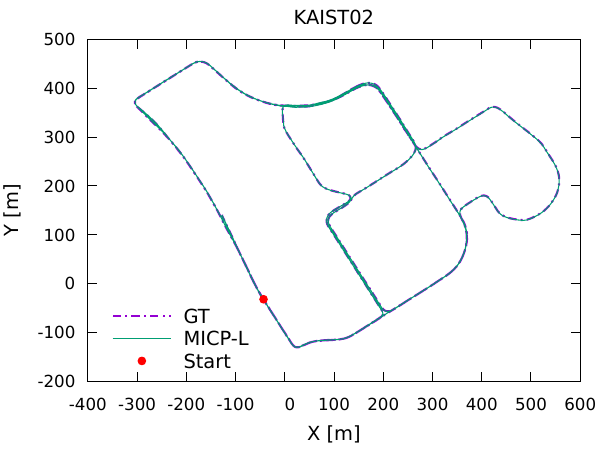}
        \includegraphics[trim={0 0 0 0},clip,width=0.325\linewidth]{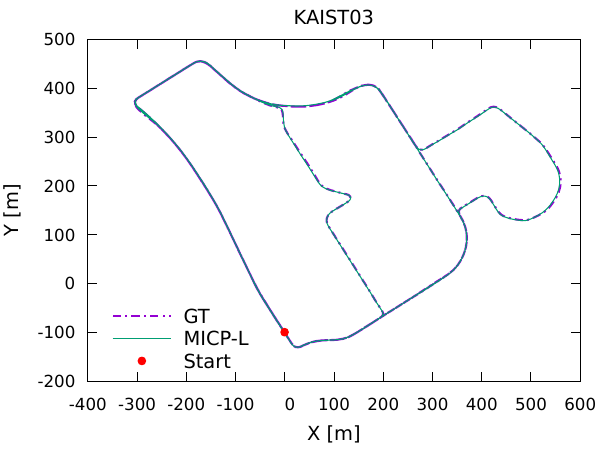}
        \vspace{-0.2cm}
    \end{subfigure}\\
    \begin{subfigure}{1.0\linewidth}
        \centering
        \includegraphics[trim={0 0 0 0},clip,width=0.245\linewidth]{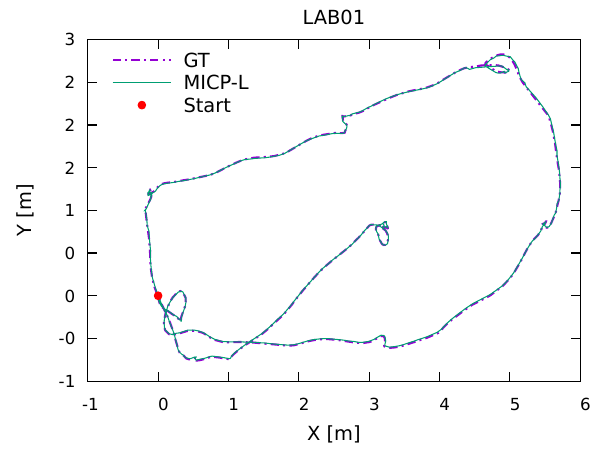}
        \includegraphics[trim={0 0 0 0},clip,width=0.245\linewidth]{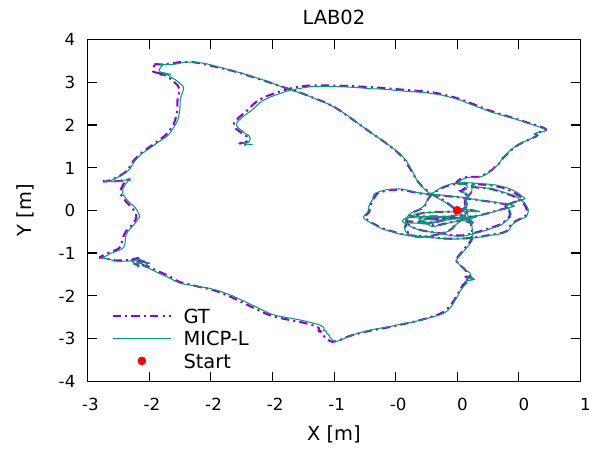}
        \includegraphics[trim={0 0 0 0},clip,width=0.245\linewidth]{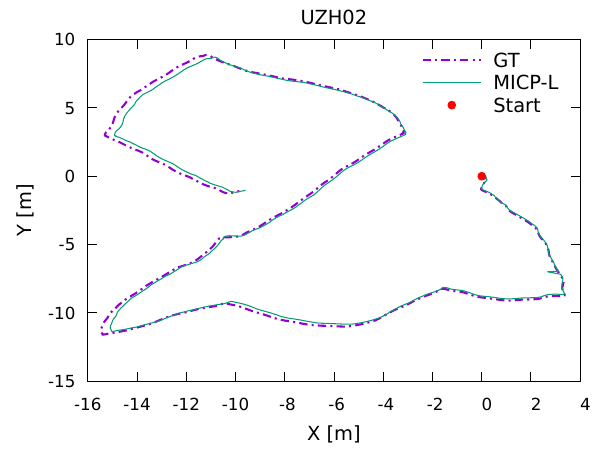}
        \includegraphics[trim={0 0 0 0},clip,width=0.245\linewidth]{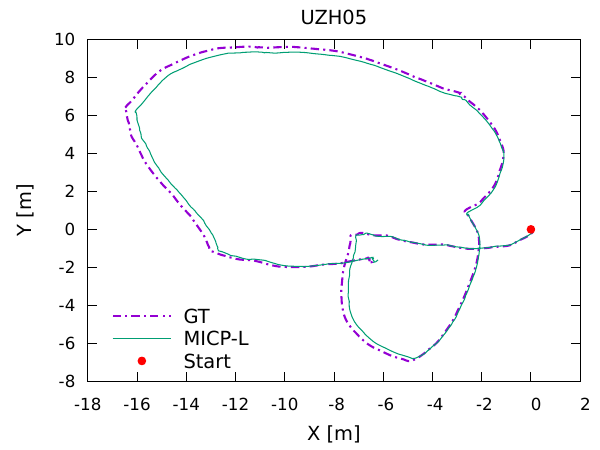}\\
        \vspace{-0.2cm}
    \end{subfigure}\\
    \caption{Trajectories estimated by MICP-L. Top row: Localization on MulRan's KAIST01, KAIST02, and KAIST03 sequence in a map generated by KAIST02 using a GPS/IMU odometry as prior. Bottom row: Localization on Hilti's LAB01, LAB02, UZH02, and UZH05 sequence in maps generated by LAB02, and UZH05, using IMU as prior. }
    \label{fig:trajectories}
    \vspace{-0.4cm}
\end{figure*}

\textbf{MulRan - Automotive}:
As datasets from the automotive domain, we selected the KAIST sequences of the MulRan datasets~\cite{kim2020mulran}.
In each of the three sequences KAIST01, KAIST02, and KAIST03 the same car drove mostly through urban areas while continuously recording data from several sensors such as a LiDAR, an IMU, and a GPS.
Moreover, each sequence is attached a single point cloud representing the entire sequence.
We used the point cloud attached to the KAIST02 sequence to reconstruct a triangle mesh using LVR2~\cite{wiemann2018irc}.
This mesh contains 6,598,646 vertices and 11,280,391 triangles covering an area of 962\,m $\times$ 710\,m.
In this single pre-generated mesh map we tracked the car's pose using all three KAIST sequences.
As simple prior odometry, we used the linear velocity from the GPS data and the orientation from IMU data.
The estimated trajectories using RCC as correspondences and P2L as optimization metric are shown in~\autoref{fig:trajectories}.
For the total travel distance of 18.37\,km we measured a mean APE between the MICP-L and ground truth trajectories of 86.6\,cm, 61.64\,cm, and 60.76\,cm for KAIST01, KAIST02, and KAIST03, respectively.
As a second metric, we tracked the rate of valid correspondences~(RVC).
We define a correspondence of a real scan point with a point on the map to be invalid if their distance is greater than 5\,m or the real scan point is outside the range measurable by the sensor.
The RVC is then the remaining number of valid points divided by the total number of points and gives a rough estimate of how many points can be considered for registration.
In outdoor domains, usually only half of the actual 3D LiDAR measurements can be used, since the other half measures towards the sky and is therefore automatically invalid.
So for these cases, the RVC would be around 50\,\% even for perfect localized robots.
On the other hand, once the RVC decreases to a lower value, it normally indicates the robot to be poorly localized.
As a third metric, we chose the point-to-mesh error on all valid correspondences (P2M) that measures how accurate a scan is matching the mesh from a given pose estimate, i.e., the actual registration performance.
The results are shown in the top part of~\autoref{tab:hilti}. 

The visualization in RViz shows that MICP-L localized the car always successfully relative to the map.
However, we observed that the map and ground truth trajectory do not match completely.
Yet, the low APE and P2M errors show that MICP-L has a high localization accuracy while constantly having low registration errors.

In addition to the ground truth baseline we executed PUMA~\cite{vizzo2021puma} with the exactly same configuration: The same map, the same distance threshold, the same point-to-plane metric using RCC, and the same odometry prior. 
However, besides not reaching the framerate of the sensor by far, PUMA lost track of the car's pose for KAIST01, KAIST02, and KAIST03 after $\sim$ 4.1\,\%, 7.4\,\%, and 12.0\,\% of the LiDAR frames, respectively.

\begin{figure}[t]
\centering
\vspace{-0.1cm}
\begin{subfigure}[t]{0.49\linewidth}
    \centering
    \includegraphics[trim=0 30 0 260,clip,width=0.99\linewidth]{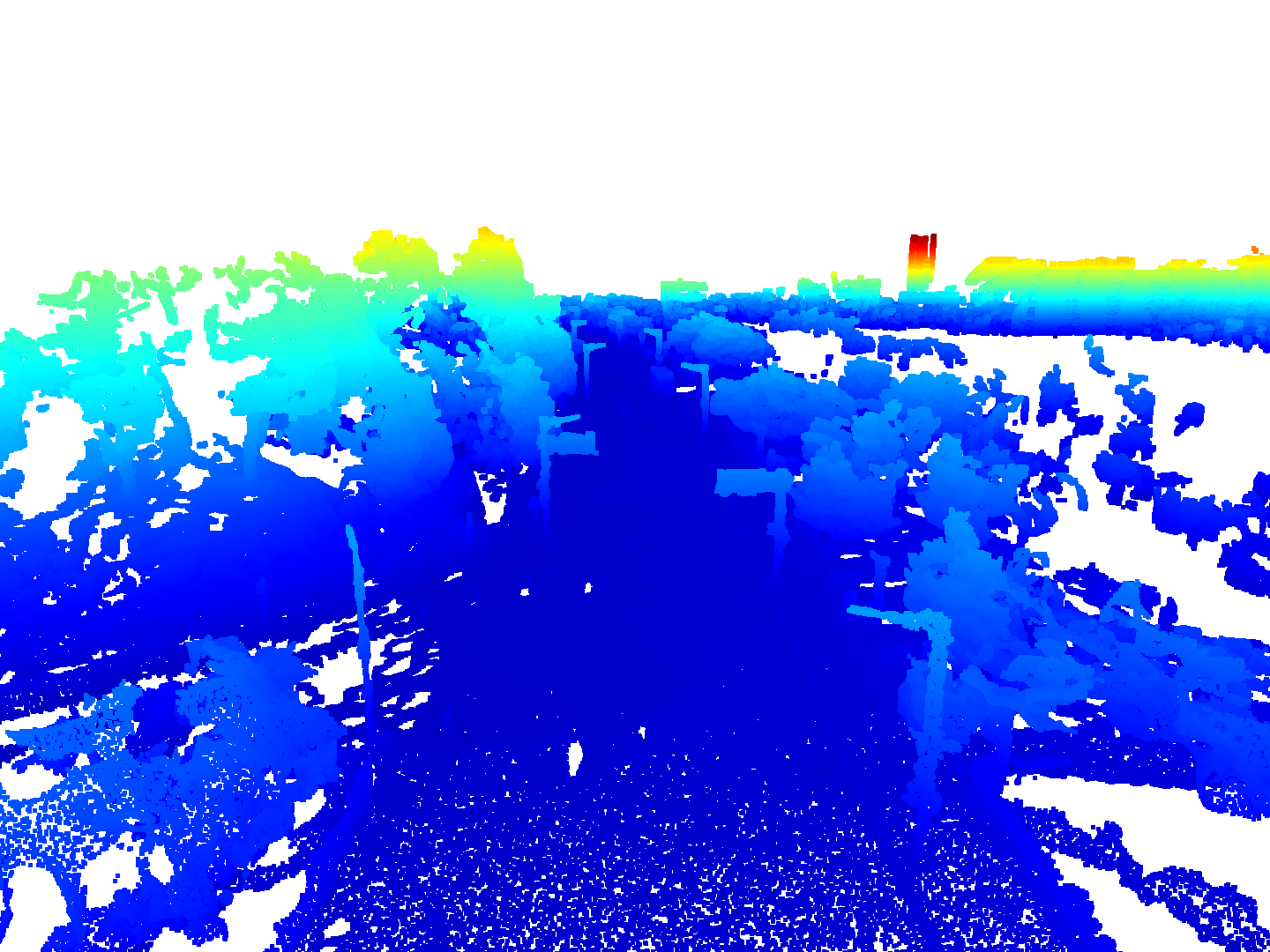}
\end{subfigure}%
\begin{subfigure}[t]{0.49\linewidth}
    \centering
    \includegraphics[trim=0 30 0 260,clip,width=0.99\linewidth]{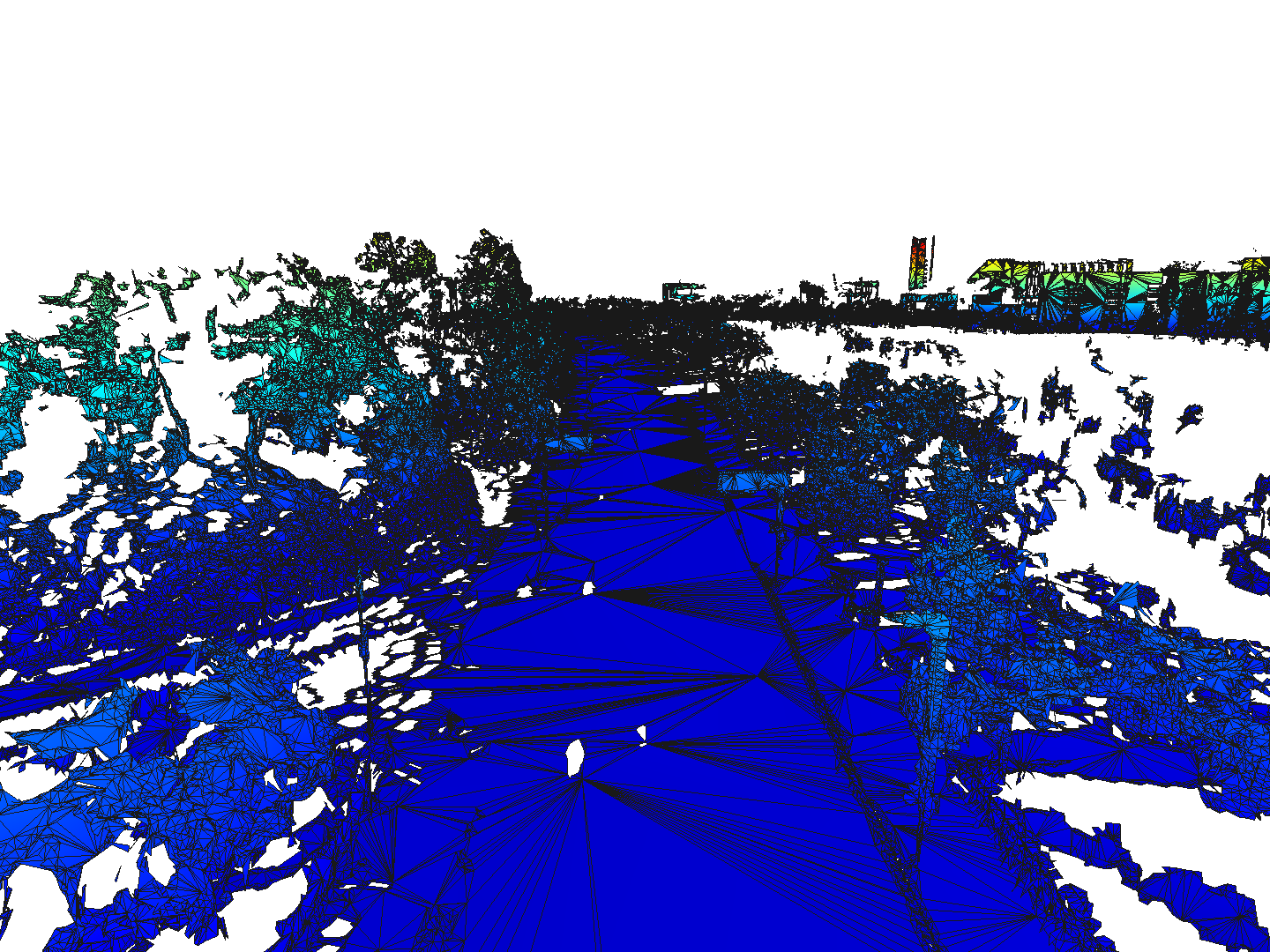}
\end{subfigure}
\caption{Maps used in the MulRan KAIST evaluation (\autoref{tab:kaist:comp}). Left: Point cloud map for KISS-ICP consisting of 33.9\,M points with a density of 100 points$/ m^3$. Right: Mesh map used with MICP-L (ours) and PUMA with 6.6\,M vertices and 11.3\,M triangles. }
\label{fig:kaist:maps}
\vspace{-0.2cm}
\end{figure}

In addition to examining mesh-based tracking algorithms, we conducted a comparative analysis with the point cloud-based method KISS-ICP~\cite{vizzo2023}.
To generate the required point cloud map, we uniformly sampled points on the already existing mesh map until reaching a density of 100 points per cubic meter, resulting in a total of 33,937,654 points in the map. 
Both the point cloud map and the mesh map are visualized in \autoref{fig:kaist:maps}.
Subsequently, after executing PUMA and KISS-ICP in tracking mode, we contrasted the outcomes with those obtained using MICP-L under various configurations.
The results, as presented in \autoref{tab:kaist:comp}, indicate that MICP-L performs on par to KISS-ICP.

\textbf{Hilti - Drone:}
In the final experiment, we used the sequences "Lab" (\emph{LAB}) 1, 2 and "RPG Drone Testing Arena" (\emph{UZH}) 2, 5 from Hilti SLAM challenge~\cite{helmberger2022hilti}, as they contain OS0-64 LiDAR data as well as 6DoF ground truth data from an external motion capture system.
We used HATSDF-SLAM~\cite{meisoldt21ecmr} to generate mesh maps from \emph{UZH02} and \emph{LAB02} using the ground truth as prior.
In \emph{LAB02}, we additionally disabled the registration in order to generate the map using the ground truth only.
With the IMU of the OS0-64 and a Madgwick filter~\cite{madgwick2010efficient}, we computed an orientation prior for MICP-L and limited the RCC to a maximum distance of 0.5\,m.
We then ran MICP-L on \emph{LAB02} and \emph{UZH02} as well as on \emph{LAB01} and \emph{UZH05}.
The estimated trajectories are shown in~\autoref{fig:trajectories}.
Again, we tracked the APE between ground truth and MICP-L trajectories.
In average, we measured errors of 2.2\,cm, 1.0\,cm, 21.5\,cm, and 22.3\,cm for LAB02, LAB01, UZH02, and UZH05, respectively.
An issue with HATSDF-SLAM's TSDF approach is that cells can be overwritten during mesh generation.
As a result, the mapping process produces accurate 3D maps but does not guarantee that the map exactly matches the ground truth, i.e., measuring the APE alone would not be meaningful.
Accordingly, we chose the RVC and P2M as second metrics to measure the actual registration performance.
The averaged P2M for both the ground truth and MICP-L are shown in the bottom part \autoref{tab:hilti}.
On Hilti datasets, all RVC scores are 100\,\% since the map is a closed surface and a scan point never exceeds the limit 5\,m from the mesh.

\begin{table}[t]
    \vspace{-0.0cm}
    \centering
    \caption{Rate of valid correspondences (RVC) and the mean point-to-mesh distance (P2M) of ground truth (GT) and MICP-L (ours) trajectory on the Hilti and MulRan datasets.}
    \label{tab:hilti}
    \begin{tabular}{ c c c c c c c }
    \toprule
     &          & \multicolumn{2}{c}{GT} & \multicolumn{2}{c}{MICP-L} \\
     & Sequence & RVC & P2M & RVC & P2M \\
     \midrule
    \multirow{3}{*}{\rotatebox[origin=c]{90}{MulRan}}
    & KAIST01 & 41.66\,\% & 48.51\,cm & \textbf{43.76\,\%} & \textbf{19.57\,cm} \\
    & KAIST02 & 41.47\,\% & 79.44\,cm & \textbf{46.94\,\%} & \textbf{16.68\,cm} \\
    & KAIST03 & 36.60\,\% & 117.31\,cm & \textbf{46.32\,\%} & \textbf{15.74\,cm} \\
    \midrule
    \multirow{4}{*}{\rotatebox[origin=c]{90}{Hilti}}
    & LAB02 & 100.0\,\% &  4.7\,cm & 100.0\,\% & \textbf{4.4\,cm} \\ 
    & LAB01 & 100.0\,\% &  6.0\,cm & 100.0\,\% & \textbf{5.2\,cm} \\
    & UZH02 & 100.0\,\% & 11.9\,cm & 100.0\,\% & \textbf{3.6\,cm} \\
    & UZH05 & 100.0\,\% & 10.1\,cm & 100.0\,\% & \textbf{3.5\,cm} \\
    \bottomrule
    \end{tabular}
    \vspace{-0.6cm}
\end{table}

\begin{table}[t]
    \vspace{-0.0cm}
    \centering
    \caption{Results of MICP-L (ours) for KAIST datasets compared to PUMA and KISS-ICP. Both MICP-L and PUMA are tracking the pose in a mesh map while KISS-ICP is using a point cloud map. Using RC or CP as correspondences and P2L or P2P as optimization. "-" marks the tracking has failed. }
    \label{tab:kaist:comp}
    \begin{tabular}{ c l r r r r r r }
    \toprule
     &          & \multicolumn{4}{c}{APE (cm)} &  map size \\
     & Algorithm & $\mu$ & $\sigma$ & min & max & (MB) \\
     \midrule
    \multirow{5}{*}{\rotatebox[origin=c]{90}{KAIST01}}
    & PUMA-RC-P2L &   -  &   -  &  -  &    -  & 214.6 \\
    & KISS-CP-P2P & 78.8 & 39.5 & 2.5 & 224.1 & 407.3 \\
    & MICP-RC-P2L & 86.6 & 37.9 & 1.8 & \textbf{211.0} & 214.6 \\
    & MICP-CP-P2L & \textbf{78.0} & 42.0 & 5.2 & 245.5 & 214.6 \\
    & MICP-CP-P2P & 81.4 & \textbf{36.9} & \textbf{1.7} & 219.8 & 214.6 \\
    \midrule
    \multirow{5}{*}{\rotatebox[origin=c]{90}{KAIST02}}
    & PUMA-RC-P2L &    - &    - &   - &     - & 214.6 \\
    & KISS-CP-P2P & 47.3 & \textbf{25.2} & 2.6 & \textbf{199.1} & 407.3 \\
    & MICP-RC-P2L & 61.6 & 37.5 & 2.4 & 237.9 & 214.6 \\
    & MICP-CP-P2L & \textbf{40.6} & 28.6 & 2.1 & 219.4 & 214.6 \\
    & MICP-CP-P2P & 48.9 & 35.5 & \textbf{1.5} & 222.1 & 214.6 \\
    \midrule
    \multirow{5}{*}{\rotatebox[origin=c]{90}{KAIST03}}
    & PUMA-RC-P2L &    - &    - &   - &     - & 214.6 \\
    & KISS-CP-P2P & 50.7 & 35.0 & 1.5 & 220.9 & 407.3 \\
    & MICP-RC-P2L & 60.8 & 39.7 & 1.2 & \textbf{213.8} & 214.6 \\
    & MICP-CP-P2L & 48.6 & 40.1 & 0.7 & 289.6 & 214.6 \\
    & MICP-CP-P2P & \textbf{42.7} & \textbf{34.9} & \textbf{0.3} & 231.1 & 214.6 \\
    \bottomrule
    \end{tabular}
    \vspace{-0.6cm}
\end{table}

For \emph{UZH} we measured high trajectory errors, but low point-to-mesh errors.
At frame 150 of \emph{UZH02} the MDE and GT P2M rises, while the MICP-L P2M remains at low errors.
Similar effects were observed in \emph{UZH05} and both \emph{LAB} sequences probably due to a map that diverges too far from the ground truth trajectory or imprecise ground truth measurements in space or time.
Therefore, the lower APEs observed when running MICP-L on the \emph{LAB} sequences are attributed to the fact that the map was generated solely from the ground truth trajectory.
This indicates that the map generated from \emph{UZH02} diverged too far from ground truth with HATSDF-SLAM.
On the \emph{LAB} sequences we also observed that the ground truth trajectories produce larger point-to-mesh errors than our approach.
The reported MICP-L P2M errors are close to the OS0-64 noise of $\pm$3\,cm.

\subsection{Convergence Study}

MICP-L implements both ray casting correspondences (RCC) and standard closest point correspondences (CPC).
The results of \autoref{tab:kaist:comp} showed that CPCs tend to produce more accurate results in the outdoors domain.
However, for most indoor scenarios, we observed that using RCCs leads to localization results more accurate and more robust to erroneous initial pose guesses.

\begin{figure}[t]
\centering
\vspace{-0.2cm}
\begin{subfigure}[t]{0.49\linewidth}
    \centering
    \includegraphics[trim=0 0 0 0,clip,width=0.99\linewidth]{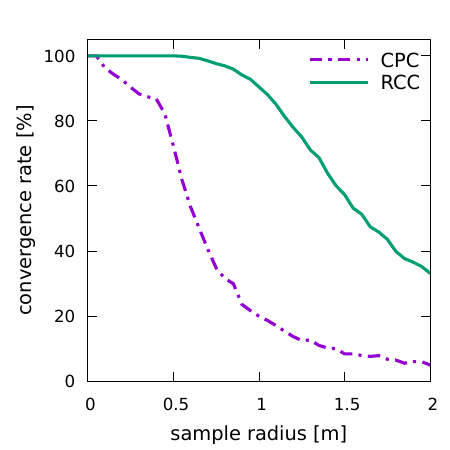}
\end{subfigure}%
\begin{subfigure}[t]{0.49\linewidth}
    \centering
    \includegraphics[trim=0 0 0 0,clip,width=0.99\linewidth]{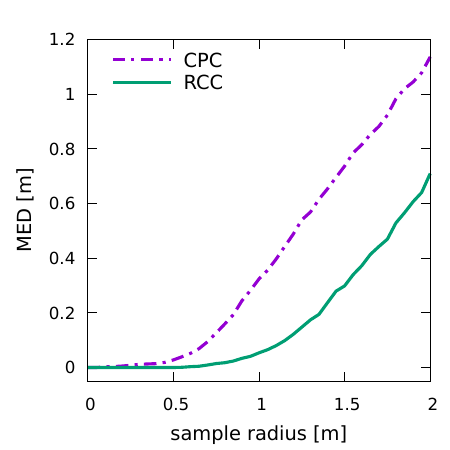}
\end{subfigure}
\caption{Convergence rates (left) and final translation errors (right) using RCC vs using CPC in AVZ map~(\autoref{fig:gazebo}). The results show that especially indoors RCC are more robust to initial guess errors than classical CPC. CPCs are used in PCL-based localization approaches as LIO-SAM~\cite{liosam2020shan} and KISS-ICP~\cite{vizzo2023}, or in non-raycasting (T)SDF-based methods like HATSDF-SLAM~\cite{meisoldt21ecmr}. }
\label{fig:convergence}
\vspace{-0.2cm}
\end{figure}

To show this, we experimented in the simulated AVZ world~\autoref{fig:gazebo} and measured the convergence rate and registration accuracy for both correspondence types.
We placed a virtual robot in the center of a room and simulated a Velodyne VLP-16 scan.
Then, we drew 512 random positions from a circular uniform distribution around the actual position of the robot.
Together with an unchanged orientation, these random positions serve as initial guesses for registration.
After 50 iterations using either RCCs or CPCs as correspondences, we measured the rate of successful convergence -- number of successful registrations divided by the total number of initial guesses -- as well as the APE of each estimated pose to the ground truth robot pose.
This experiment was repeated in every room available in the AVZ world.
We tracked the APE and convergence rates while increasing the radius of the uniform sample distribution.
The averaged results over all rooms are shown in~\autoref{fig:convergence}.

The results confirm our initial thesis.
Especially when we observed the convergences of the CPC, it often happened that the scan was matched to the opposite side of the wall. 
Then optimization reached a local minimum and ended.
In contrast, with ray casting we automatically find the nearest surface as correspondence, thus, it is more likely to be "next" to the robot.

\subsection{Computational Demands}

\begin{table}[t]
    \centering
    \caption{Computers used for run time evaluation.}
    \label{tab:devices}
    \begin{tabular}{ c c c }
    \toprule
    Name & CPU & GPU \\
    \midrule
    DPC & AMD Ryzen 7 3700X & NVIDIA RTX 2070 SUPER \\ 
    NUC & Intel i7 1165G7 & NVIDIA RTX 2060 Max-P \\ 
    LX1 & Intel i7 8750H & NVIDIA GTX 1050 Ti Max-Q \\
    \bottomrule
    \end{tabular}
    \vspace{-0.4cm}
\end{table}

\begin{figure}[t]
    \centering
    \vspace{-0.2cm}
    \includegraphics[width=0.48\linewidth]{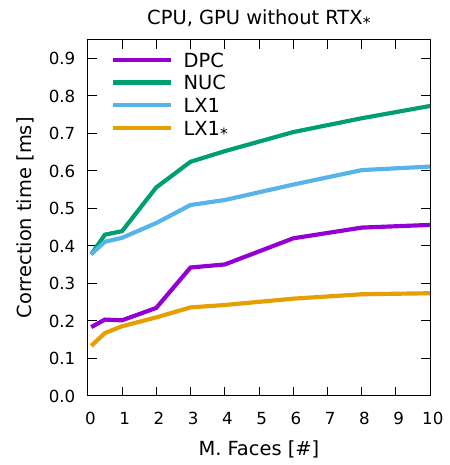}
    \includegraphics[width=0.48\linewidth]{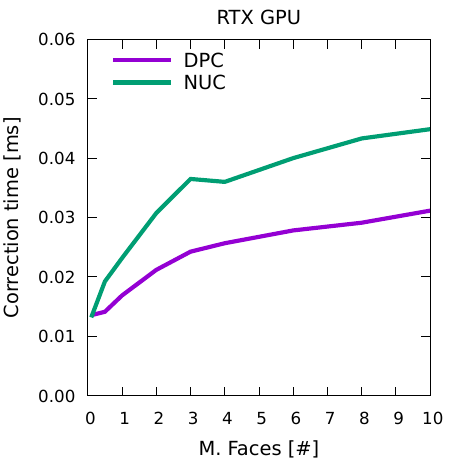}
    \caption{Run time evaluation of a single MICP-L iteration in synthetic maps of varying sizes on different architectures for CPUs and GPUs (left) and accelerated with RTX hardware (right).}
    \label{fig:runtime}
\end{figure}

First, we analyze the run time of our GPU and CPU implementation, considering varying map sizes, and different computing devices.
The computers used in experiments are listed in \autoref{tab:devices}.
For each of the experiments shown in \autoref{fig:runtime}, we measured the run time for a Velodyne VLP-16 LiDAR as a virtual sensor.
The VLP-16 has 16 scan lines and operates with frame rates up to 20 Hz.
In the horizontal direction, the number of points is adjustable.
In our experiments, we used 900 horizontal points, requiring the calculation of 14,400 RCCs.
Next, we spawned 1,000 random pose guesses inside an automatically generated sphere and simulated a VLP-16 scan at the sphere's center.
Afterwards, we used MICP-L to register this scan from each pose estimate against the sphere, so that every random pose estimate converges at the center of the sphere.
We measured the run times per correction step for different numbers of sphere triangles on all devices listed in~\autoref{tab:devices}.
The results are summarized in \autoref{fig:runtime}.
The experiments show that our proposed algorithm is online capable in 3D maps consisting of many triangles.
With the Intel NUC's RTX hardware-acceleration, a VLP-16 scan is corrected about 850,000 times in a map of 1 million (M) faces between two scans.

In a second experiment, we measured each component of MICP-L (\autoref{fig:workflow}) for all considered devices by spawning a sphere consisting of 1 million (M) faces and registering 1,000 poses simultaneously against the mesh.
The results in~\autoref{tab:runtimesinner} provide an overview of how much time each stage of the correction needs in relation to the total computation time.
They show by utilizing more specialized hardware, the bottleneck associated with complex ray casting calculations becomes less significant.

The experiments were conducted using purely synthetic data; however, the findings are directly applicable to real data.
This is because the runtime mostly depends on finding the RCCs which is not affected by the noise of a real sensor.

\begin{table}[t]
    \centering
    \vspace{-0.1cm}
    \caption{Partial relative run times of each MICP-L step using CPU, non-RTX-GPU (GPU), and RTX-GPU (RTX) for correcting VLP-16 scans in a map of 1 million triangles. }
    \label{tab:runtimesinner}
    \begin{tabular}{ c r r r r }
    \toprule
    Device & FindRCC & Reduction & SVD \\
    \midrule 
    CPU & 96.40\,\% & 3.59\,\% & 0.01\,\% \\
    GPU & 94.09\,\% & 0.01\,\% & 5.90\,\% \\
    RTX & 80.33\,\% & 0.38\,\% & 19.29\,\% \\
    \bottomrule
    \end{tabular}
    \vspace{-0.4cm}
\end{table}

\section{Conclusion}

We presented MICP-L, an approach to robustly localize a robot equipped with one or more range sensors in triangle mesh maps of large scale.
MICP-L is specifically designed for robotic applications, due to its compatibility with arbitrary combinations of range sensors.
The required computational steps can be flexibly distributed to both CPU and GPU, making our method adaptable to diverse robotic platforms and target architectures.
Our method was evaluated in terms of localization performance across the domains of agriculture, drones, and automotive applications.
We showed that our method is on par with KISS-ICP and more reliable than PUMA for outdoors domains.
In addition, with our hardware-accelerated approach of computing ray casting correspondences, we showed significant advantages over classic closest point correspondences in terms of runtime, convergence rate and registration accuracy for indoor domains.
MICP-L has been successfully integrated into a state-of-the-art mesh navigation software that enables AGVs to navigate safely and autonomously in a variety of scenarios.

In our forthcoming research, we will focus on the global localization problem through a multimodal approach to state estimation.
Moreover, we have identified that our localization method automatically distinguishes range measurements into map and non-map points. 
We intend to utilize this capability as a preprocessing stage to accelerate and enhance obstacle detection algorithms.

\bibliographystyle{IEEEtran}
\bibliography{IEEEabrv, references}

\begin{thebibliography}{10}
\providecommand{\url}[1]{#1}
\csname url@rmstyle\endcsname
\providecommand{\newblock}{\relax}
\providecommand{\bibinfo}[2]{#2}
\providecommand\BIBentrySTDinterwordspacing{\spaceskip=0pt\relax}
\providecommand\BIBentryALTinterwordstretchfactor{4}
\providecommand\BIBentryALTinterwordspacing{\spaceskip=\fontdimen2\font plus
\BIBentryALTinterwordstretchfactor\fontdimen3\font minus
  \fontdimen4\font\relax}
\providecommand\BIBforeignlanguage[2]{{%
\expandafter\ifx\csname l@#1\endcsname\relax
\typeout{** WARNING: IEEEtran.bst: No hyphenation pattern has been}%
\typeout{** loaded for the language `#1'. Using the pattern for}%
\typeout{** the default language instead.}%
\else
\language=\csname l@#1\endcsname
\fi
#2}}

\bibitem{meisoldt21ecmr}
M.~Eisoldt, M.~Flottmann, J.~Gaal, P.~Buschermöhle, S.~Hinderink, M.~Hillmann,
  A.~Nitschmann, P.~Hoffmann, T.~Wiemann, and M.~Porrmann, ``{HATSDF SLAM –
  Hardware-accelerated TSDF SLAM for Reconfigurable SoCs},'' in \emph{European
  Conference on Mobile Robots (ECMR)}.\hskip 1em plus 0.5em minus 0.4em\relax
  IEEE, 2021, pp. 1--7.

\bibitem{puetz2021cvp}
S.~Pütz, T.~Wiemann, M.~Kleine~Piening, and J.~Hertzberg, ``{Continuous
  Shortest Path Vector Field Navigation on 3D Triangular Meshes for Mobile
  Robots},'' in \emph{International Conference on Robotics and Automation
  (ICRA)}.\hskip 1em plus 0.5em minus 0.4em\relax IEEE, 2021, pp. 2256--2263.

\bibitem{mock2023rmagine}
A.~Mock, T.~Wiemann, and J.~Hertzberg, ``{Rmagine: 3D Range Sensor Simulation
  in Polygonal Maps via Raytracing for Embedded Hardware on Mobile Robots},''
  in \emph{International Conference on Robotics and Automation (ICRA)}.\hskip
  1em plus 0.5em minus 0.4em\relax IEEE, 2023, pp. 9076--9082.

\bibitem{shan2018lego}
T.~Shan and B.~Englot, ``{LeGO-LOAM: Lightweight and Ground-Optimized Lidar
  Odometry and Mapping on Variable Terrain},'' in \emph{International
  Conference on Intelligent Robots and Systems (IROS)}.\hskip 1em plus 0.5em
  minus 0.4em\relax IEEE, 2018, pp. 4758--4765.

\bibitem{liosam2020shan}
T.~Shan, B.~Englot, D.~Meyers, W.~Wang, C.~Ratti, and R.~Daniela, ``{LIO-SAM:
  Tightly-coupled Lidar Inertial Odometry via Smoothing and Mapping},'' in
  \emph{International Conference on Intelligent Robots and Systems
  (IROS)}.\hskip 1em plus 0.5em minus 0.4em\relax IEEE, 2020, pp. 5135--5142.

\bibitem{fastlio2021}
W.~Xu and F.~Zhang, ``{FAST-LIO: A Fast, Robust LiDAR-Inertial Odometry Package
  by Tightly-Coupled Iterated Kalman Filter},'' \emph{IEEE Robotics and
  Automation Letters (RAL)}, vol.~6, no.~2, pp. 3317--3324, 2021.

\bibitem{fasterlio2022}
C.~Bai, T.~Xiao, Y.~Chen, H.~Wang, F.~Zhang, and X.~Gao, ``{Faster-LIO:
  Lightweight Tightly Coupled Lidar-Inertial Odometry Using Parallel Sparse
  Incremental Voxels},'' \emph{IEEE Robotics and Automation Letters (RAL)},
  vol.~7, no.~2, pp. 4861--4868, 2022.

\bibitem{vizzo2023}
I.~Vizzo, T.~Guadagnino, B.~Mersch, L.~Wiesmann, J.~Behley, and C.~Stachniss,
  ``{KISS-ICP: In Defense of Point-to-Point ICP – Simple, Accurate, and
  Robust Registration If Done the Right Way},'' \emph{IEEE Robotics and
  Automation Letters (RAL)}, vol.~8, no.~2, pp. 1029--1036, 2023.

\bibitem{oleynikova2017voxblox}
H.~Oleynikova, Z.~Taylor, M.~Fehr, R.~Siegwart, and J.~Nieto, ``{Voxblox:
  Incremental 3D Euclidean Signed Distance Fields for On-Board MAV Planning},''
  in \emph{International Conference on Intelligent Robots and Systems
  (IROS)}.\hskip 1em plus 0.5em minus 0.4em\relax IEEE, 2017, pp. 1366--1373.

\bibitem{vizzo2022sensors}
\BIBentryALTinterwordspacing
I.~Vizzo, T.~Guadagnino, J.~Behley, and C.~Stachniss, ``{VDBFusion: Flexible
  and Efficient TSDF Integration of Range Sensor Data},'' \emph{Sensors},
  vol.~22, no.~3, 2022. [Online]. Available:
  \url{https://www.mdpi.com/1424-8220/22/3/1296}
\BIBentrySTDinterwordspacing

\bibitem{vizzo2021puma}
I.~Vizzo, X.~Chen, N.~Chebrolu, J.~Behley, and C.~Stachniss, ``{Poisson Surface
  Reconstruction for LiDAR Odometry and Mapping},'' in \emph{International
  Conference on Robotics and Automation (ICRA)}.\hskip 1em plus 0.5em minus
  0.4em\relax IEEE, 2021, pp. 5624--5630.

\bibitem{dai2017bundlefusion}
A.~Dai, M.~Nie{\ss}ner, M.~Zollh{\"o}fer, S.~Izadi, and C.~Theobalt,
  ``{BundleFusion: Real-Time Globally Consistent 3D Reconstruction Using
  On-the-Fly Surface Reintegration},'' \emph{ACM Transactions on Graphics
  (ToG)}, vol.~36, no.~4, p.~1, 2017.

\bibitem{akai2020}
N.~Akai, T.~Hirayama, and H.~Murase, ``{3D Monte Carlo Localization with
  Efficient Distance Field Representation for Automated Driving in Dynamic
  Environments},'' in \emph{Intelligent Vehicles Symposium (IV)}.\hskip 1em
  plus 0.5em minus 0.4em\relax IEEE, 2020, pp. 1859--1866.

\bibitem{dreher2021global}
M.~Dreher, H.~Blum, R.~Siegwart, and A.~Gawel, ``{Global Localization in
  Meshes},'' in \emph{International Symposium on Automation and Robotics in
  Construction (ISARC)}, vol.~38.\hskip 1em plus 0.5em minus 0.4em\relax IAARC,
  2021, pp. 747--754.

\bibitem{chen2021range}
X.~Chen, I.~Vizzo, T.~L{\"a}be, J.~Behley, and C.~Stachniss, ``{Range
  Image-based LiDAR Localization for Autonomous Vehicles},'' in
  \emph{International Conference on Robotics and Automation (ICRA)}.\hskip 1em
  plus 0.5em minus 0.4em\relax IEEE, 2021, pp. 5802--5808.

\bibitem{ruan2023slamesh}
J.~Ruan, B.~Li, Y.~Wang, and Y.~Sun, ``{SLAMesh: Real-time LiDAR Simultaneous
  Localization and Meshing},'' in \emph{International Conference on Robotics
  and Automation (ICRA)}.\hskip 1em plus 0.5em minus 0.4em\relax IEEE, 2023,
  pp. 3546--3552.

\bibitem{zhong2023}
X.~Zhong, Y.~Pan, J.~Behley, and C.~Stachniss, ``{SHINE-Mapping: Large-Scale 3D
  Mapping Using Sparse Hierarchical Implicit Neural Representations},'' in
  \emph{International Conference on Robotics and Automation (ICRA)}.\hskip 1em
  plus 0.5em minus 0.4em\relax IEEE, 2023, pp. 8371--8377.

\bibitem{wiesmann2023}
L.~Wiesmann, T.~Guadagnino, I.~Vizzo, N.~Zimmerman, Y.~Pan, H.~Kuang,
  J.~Behley, and C.~Stachniss, ``{LocNDF: Neural Distance Field Mapping for
  Robot Localization},'' \emph{IEEE Robotics and Automation Letters (RAL)},
  vol.~8, no.~8, pp. 4999--5006, 2023.

\bibitem{wiemann2018irc}
T.~Wiemann, I.~Mitschke, A.~Mock, and J.~Hertzberg, ``{Surface Reconstruction
  from Arbitrarily Large Point Clouds},'' in \emph{International Conference on
  Robotic Computing (IRC)}.\hskip 1em plus 0.5em minus 0.4em\relax IEEE, 2018,
  pp. 278--281.

\bibitem{lin2023immesh}
J.~Lin, C.~Yuan, Y.~Cai, H.~Li, Y.~Ren, Y.~Zou, X.~Hong, and F.~Zhang,
  ``{ImMesh: An Immediate LiDAR Localization and Meshing Framework},''
  \emph{IEEE Transactions on Robotics}, vol.~39, no.~6, pp. 4312--4331, 2023.

\bibitem{rusinkiewicz2001efficient}
S.~Rusinkiewicz and M.~Levoy, ``{Efficient Variants of the ICP Algorithm},'' in
  \emph{International Conference on 3-D Digital Imaging and Modeling
  (3DIM)}.\hskip 1em plus 0.5em minus 0.4em\relax IEEE, 2001, pp. 145--152.

\bibitem{koide2021vgicp}
K.~Koide, M.~Yokozuka, S.~Oishi, and A.~Banno, ``{Voxelized GICP for Fast and
  Accurate 3D Point Cloud Registration},'' in \emph{International Conference on
  Robotics and Automation (ICRA)}.\hskip 1em plus 0.5em minus 0.4em\relax IEEE,
  2021, pp. 11\,054--11\,059.

\bibitem{segal2009gicp}
A.~Segal, D.~Haehnel, and S.~Thrun, ``{Generalized-ICP},'' in \emph{Robotics:
  science and systems (RSS)}, 2009.

\bibitem{holz2014registration}
D.~Holz and S.~Behnke, ``{Registration of Non-Uniform Density 3D Point Clouds
  using Approximate Surface Reconstruction},'' in \emph{International Symposium
  on Robotics (ISR)}.\hskip 1em plus 0.5em minus 0.4em\relax VDE, 2014, pp.
  1--7.

\bibitem{li2015modified}
W.~Li and P.~Song, ``{A modified ICP algorithm based on dynamic adjustment
  factor for registration of point cloud and CAD model},'' \emph{Pattern
  Recognition Letters}, vol.~65, pp. 88--94, 2015.

\bibitem{avetisyan2019scan2cad}
A.~Avetisyan, M.~Dahnert, A.~Dai, M.~Savva, A.~X. Chang, and M.~Nie{\ss}ner,
  ``{Scan2CAD: Learning CAD Model Alignment in RGB-D Scans},'' in
  \emph{Conference on Computer Vision and Pattern Recognition (CVPR)}.\hskip
  1em plus 0.5em minus 0.4em\relax IEEE, 2019, pp. 2614--2623.

\bibitem{mejia2019pcl2mesh}
D.~Mejia-Parra, J.~Lalinde-Pulido, J.~R. Sánchez, O.~Ruiz-Salguero, and
  J.~Posada, ``{Perfect Spatial Hashing for Point-cloud-to-mesh
  Registration},'' in \emph{CEIG}.\hskip 1em plus 0.5em minus 0.4em\relax The
  Eurographics Association, 2019, pp. 41--50.

\bibitem{bourquat2022hierarchical}
P.~Bourquat, D.~Coeurjolly, G.~Damiand, and F.~Dupont, ``{Hierarchical
  mesh-to-points as-rigid-as-possible registration},'' \emph{Computers \&
  Graphics}, vol. 102, pp. 320--328, 2022.

\bibitem{zhang2022icp}
J.~Zhang, Y.~Yao, and B.~Deng, ``{Fast and Robust Iterative Closest Point},''
  \emph{IEEE Transactions on Pattern Analysis and Machine Intelligence},
  vol.~44, no.~7, pp. 3450--3466, 2022.

\bibitem{embree2014}
I.~Wald, S.~Woop, C.~Benthin, G.~S. Johnson, and M.~Ernst, ``{Embree: A Kernel
  Framework for Efficient CPU Ray Tracing},'' \emph{ACM Transactions on
  Graphics (TOG)}, 2014.

\bibitem{optix2010}
S.~G. Parker, J.~Bigler, A.~Dietrich, H.~Friedrich, J.~Hoberock, D.~Luebke,
  D.~McAllister, M.~McGuire, K.~Morley, A.~Robison, and M.~Stich, ``{OptiX: A
  General Purpose Ray Tracing Engine},'' \emph{ACM Transactions on Graphics
  (TOG)}, 2010.

\bibitem{umeyama1991}
S.~Umeyama, ``{Least-squares estimation of transformation parameters between
  two point patterns},'' \emph{IEEE Pattern Analysis \& Machine Intelligence
  (PAMI)}, vol.~13, no.~04, pp. 376--380, 1991.

\bibitem{puetz2018mbf}
S.~Pütz, J.~S. Simón, and J.~Hertzberg, ``{Move Base Flex: A Highly Flexible
  Navigation Framework for Mobile Robots},'' in \emph{International Conference
  on Intelligent Robots and Systems (IROS)}.\hskip 1em plus 0.5em minus
  0.4em\relax IEEE, 2018, pp. 3416--3421.

\bibitem{kim2020mulran}
G.~{Kim}, Y.~S. {Park}, Y.~{Cho}, J.~{Jeong}, and A.~{Kim}, ``{MulRan:
  Multimodal Range Dataset for Urban Place Recognition},'' in
  \emph{International Conference on Robotics and Automation (ICRA)}.\hskip 1em
  plus 0.5em minus 0.4em\relax IEEE, 2020, pp. 6246--6253.

\bibitem{helmberger2022hilti}
M.~Helmberger, K.~Morin, B.~Berner, N.~Kumar, G.~Cioffi, and D.~Scaramuzza,
  ``{The Hilti SLAM Challenge Dataset},'' \emph{IEEE Robotics and Automation
  Letters (RAL)}, vol.~7, no.~3, pp. 7518--7525, 2022.

\bibitem{madgwick2010efficient}
S.~Madgwick \emph{et~al.}, ``{An efficient orientation filter for inertial and
  inertial/magnetic sensor arrays},'' \emph{Report x-io and University of
  Bristol (UK)}, 2010.

\end{thebibliography}

\end{document}